\documentclass{article} 
\usepackage{iclr2026_conference,times}

\usepackage{amsmath,amsfonts,bm}

\def\eqref#1{equation~\ref{#1}}

\def\1{\bm{1}}

\DeclareMathAlphabet{\mathsfit}{\encodingdefault}{\sfdefault}{m}{sl}
\SetMathAlphabet{\mathsfit}{bold}{\encodingdefault}{\sfdefault}{bx}{n}

\usepackage{hyperref}
\usepackage{url}

\usepackage[utf8]{inputenc} 
\usepackage[T1]{fontenc}    
\usepackage{hyperref}       
\usepackage{url}            
\usepackage{booktabs}       
\usepackage{amsfonts}       
\usepackage{nicefrac}       
\usepackage{microtype}      
\usepackage{xcolor}         

\usepackage{graphicx}
\usepackage{natbib}
\usepackage{amsfonts}       
\usepackage{enumitem}       
\usepackage{subcaption}
\usepackage{booktabs} 
\usepackage{tabularx} 
\usepackage{wrapfig}
\usepackage{amssymb}
\usepackage{amsmath}
\usepackage{marvosym}
\usepackage{pdfpages}

\usepackage{tikz}
\usepackage{mathdots}
\usepackage{yhmath}
\usepackage{cancel}
\usepackage{color}
\usepackage{siunitx}
\usepackage{array}
\usepackage{multirow}
\usepackage{amssymb}
\usepackage{gensymb}
\usepackage{tabularx}
\usepackage{extarrows}
\usetikzlibrary{fadings}
\usetikzlibrary{patterns}
\usetikzlibrary{shadows.blur}
\usetikzlibrary{shapes}
\usepackage{colortbl} 
\usepackage{xcolor}   

\usepackage{tikz}
\usetikzlibrary{positioning, arrows.meta}

\definecolor{cividisBluePositive}{HTML}{005A9C} 
\definecolor{neutralGreyNegative}{HTML}{686868} 

\title{Analyzing Generalization in Pre-Trained\\ Symbolic Regression}

\author{Henrik Voigt, Paul Kahlmeyer, Kai Lawonn, Michael Habeck \& Joachim Giesen   \\
Department of Computer Science\\
University of Jena\\
Jena, Germany \\
\texttt{\{first.last\}@uni-jena.de} \\
\vspace{-2.5em}
}

\iclrfinalcopy 

\usepackage{fancyhdr}
\begin{document}

\maketitle

\begin{abstract}
Symbolic regression algorithms search a space of mathematical expressions for formulas that explain given data. 
Transformer-based models have emerged as a promising, scalable approach 
shifting the expensive combinatorial search to a large-scale pre-training phase. 
However, the success of these models is critically dependent on their pre-training data. 
Their ability to generalize to problems outside of this pre-training distribution remains largely unexplored. 
In this work, we conduct a systematic empirical study to evaluate the generalization capabilities of pre-trained, transformer-based symbolic regression. 
We rigorously test performance both within the pre-training distribution and on a series of out-of-distribution challenges for several state of the art approaches. 
Our findings reveal a significant dichotomy: while pre-trained models perform well in-distribution, the performance consistently degrades in out-of-distribution scenarios. 
We conclude that this generalization gap is a critical barrier for practitioners, as it severely limits the practical use of pre-trained approaches for real-world applications. 
\end{abstract}

\section{Introduction}
\label{sec:Introduction}
The traditional objective in regression is generalization, that is, learning a function from training data that performs well beyond the training data.
Symbolic regression adds another objective, namely, interpretability of the regression function.
Therefore, symbolic regression has found applications in almost all areas of the natural sciences~\citep{udrescu2021symbolic,cornelio2023combining,camps2023discovering}, engineering~\citep{wu2023enhancing,abdusalamov2023automatic,tsoi2024symbolic}, and medicine~\citep{christensen2022identifying,la2023flexible,zhang2024discovering}.
Symbolic regression algorithms are typically evaluated along three dimensions~\citep{srbench_lacava21}: 
their accuracy on held-out data, the complexity of their description, and their ability to recover ground truth formulas from the training data.
The first dimension, \emph{accuracy}, is a classical measure of generalization ability, and the second dimension, \emph{complexity}, is a measure of interpretability.
The third dimension, \emph{recovery}, is a strong measure of both, interpretability and generalization ability.

In terms of recovery, the state of the art in symbolic regression is achieved by search-based 
algorithms.
Since the search space of symbolic expressions grows exponentially with the complexity of the expressions, search-based symbolic regression incurs high training costs when searching for a symbolic regression function. 
This motivated the recently proposed transformer-based approach, which shifts a large part of the cost to a pre-training phase. This new paradigm, however, introduces a dependency on the pre-training data—typically a vast collection of synthetic mathematical expressions. The success of these models implicitly relies on the assumption that they learn robust, generalizable representations. Yet, this assumption remains largely untested. This raises a crucial question: 
\textit{To what extend do pre-trained symbolic regression approaches generalize?}



\begin{figure*}[t!]
  \centering  
  \includegraphics[width=0.999\textwidth]{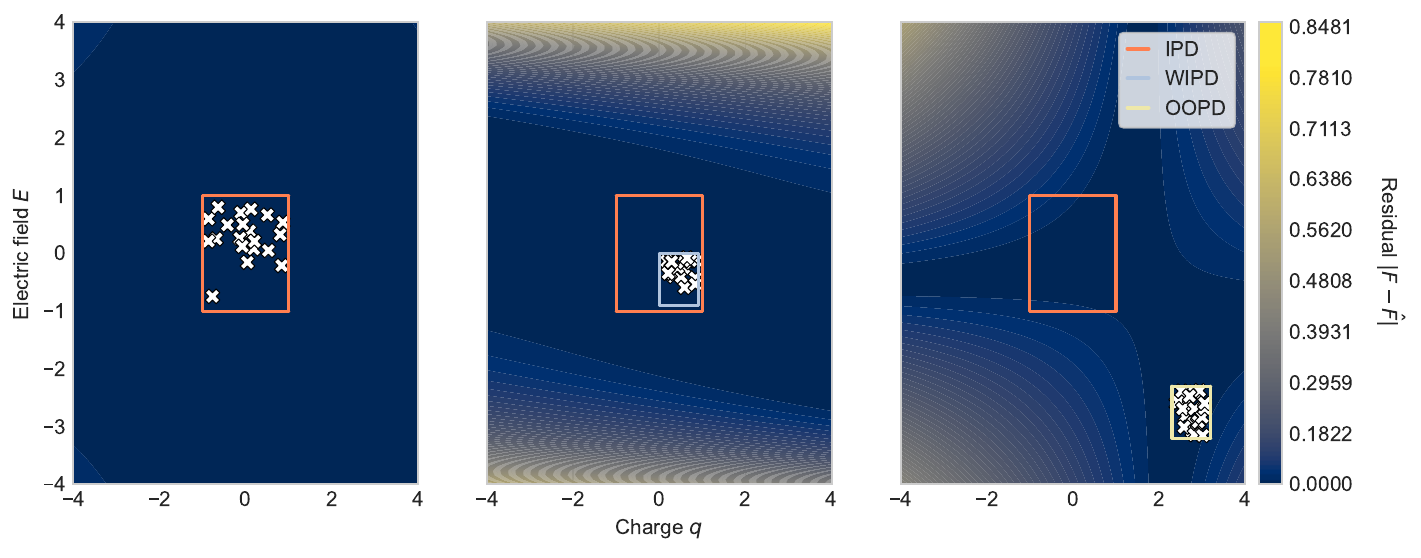} 
\caption{\textbf{Performance of transformer-based regression methods across domains.} The figure illustrates the performance of the transformer-based regression method E2E~\citep{kamienny22_transformer} when applied to training data from different domains, using the Coulomb force formula ($F=q\cdot E$) as a case study. The training data (white crosses) are sampled from the pre-training domain (IPD, red), a subdomain of the pre-training domain (WIPD, blue), and a domain disjoint from the pre-training domain (OOPD, yellow). The regression method generates formulas $\hat F$: $0.998q\cdot E$ (IPD), $1.01q\cdot E + 0.002 q^4 + 0.002 E^3$ (WIPD), and $1.02q\cdot E + 0.014q - 0.039E$ (OOPD). While the WIPD and OOPD expressions deviate from the ground truth, they remain relatively accurate near the training data, as indicated by low residual values.}
\label{fig:intro_example}
\vspace{-1.5em}
\end{figure*}

In Figure~\ref{fig:intro_example}, we demonstrate the effect of three training data sampling regimes— in-pretraining domain (IPD), within-pre-training domain (WIPD), and out-of-pre-training domain (OOPD)—on the 
performance of the transformer-based symbolic regression algorithm E2E~\citep{kamienny22_transformer}. We use the simple example of a Coulomb force $F$ acting on a charge $q$ in an electric field $E$, which is given as $F=q\cdot E$. We observe that for IPD and WIPD data, both the model's predictions and generated symbolic expressions remain close to the ground truth. In contrast, on OOPD data, the increased residuals and deviating symbolic form indicate challenges with OOPD generalization.

In this work, we conduct the first systematic empirical study to analyze the generalization capabilities of pre-trained, transformer-based symbolic regression approaches. We compare five representative approaches against a search-based baseline using the SRBench benchmark~\citep{srbench_lacava21}. We investigate their performance when applied to problems that are in-, within- and out-of- their pre-training distribution. 
Additionally, we evaluate the robustness to a suite of common data perturbations—including input domain scaling and shifting, distribution changes, and output noise—that are frequently encountered in practical applications.

Our contributions are as follows:

\begin{enumerate}
[
    label=\textbullet,
    topsep=0pt,         
    itemsep=4pt,        
    partopsep=0pt,      
    parsep=0pt,
    leftmargin=12pt
]
    \item We provide a rigorous evaluation of leading transformer-based symbolic regression approaches, revealing a significant performance gap: while they perform well on problems in and within their pre-training distributions, their performance significantly degrades in out-of-pre-training scenarios.    
    \item We demonstrate the fragility of pre-trained models by showing that their performance deteriorates when subjected to common data perturbations, such as input shifts, scaling and noise.
    \item We show that, as a consequence of this limitation, current pre-trained symbolic regression approaches are not competitive with a search-based baseline on problems that require generalization.
    \item We provide crucial insights for practitioners, highlighting the risks of applying these models to real-world data that may not match the pre-training distribution, and suggest future research directions to build more robust models.
\end{enumerate}

\section{Related Work}
\label{sec:related_work}
\textbf{From search to pre-training in symbolic regression.} State-of-the-art symbolic regression methods either implicitly~\citep{dsr4_petersen21} or explicitly~\citep{pysr_cranmer23, kahlmeyer2024udfs} search some space of small expressions that are built from a predefined set of operators, usually represented by expression trees.
Although symbolic regression aims for small expressions, because large expressions lose interpretability, the search cost can be substantial. This is because the space of expression trees grows exponentially with the number of variable and operator nodes in the tree~\citep{virgolin2022symbolic}. 
Therefore, \citet{biggio2021neural} proposed, in seminal work, scalable neural symbolic regression with transformers to mitigate the computationally intensive search problem by shifting the heavy lifting into a pre-training phase. 
Recently, the transformer-approach to symbolic regression has seen a surge in popularity~\citep{kamienny22_transformer, lalande23_transformer, Li2023TransformerbasedMF, li2025discovering, merler2024context, grayeli2024symbolic, shojaee2024llm, shojaee2025llm, vastl2024symformer}.
Common to all these works is the use of the transformer architecture~\citep{VaswaniSPUJGKP17} for translating sequences of sample points into sequences of tokens that represent formulas.
They differ in the encoding of the sample points, the representation of the formulas, and in the choice of loss functions.
Moreover, constants in formula skeletons are either fitted by numerical optimization as in~\citep{biggio2021neural}, or, are predicted, in an end-to-end approach, as part of the formulas~\citep{kamienny22_transformer}.

\textbf{The central role of pre-training data.} A key distinction of all these implementations of the transformer approach from search-based methods is that the performance of transformer-based methods is largely contingent upon their pre-training data. Consequently, a significant body of work has focused on optimizing the generation of this pre-training data, which involves two main axes: (1) selecting a representative 
set of symbolic formulas, and (2) defining a strategy to sample numerical data points from these functions.

\subsection{Selection of Symbolic Expressions}
\label{sec:expression_selection}
\textbf{Expression tree sampling.} A common pre-training strategy is to select symbolic formulas from a large corpus of randomly generated expression trees. 
\citet{biggio2021neural}, \citet{kamienny22_transformer}, and \citet{vastl2024symformer} utilize the framework of \citet{lample2019deep} for randomly generating these trees. 
Within this framework, a symbolic formula is represented as a tree in which the internal nodes are labeled with unary and binary operators, and the leaf nodes are labeled with variables and constants. 
The random generation strategy, however, presents notable drawbacks.
First, sampled symbolic expressions that are syntactically different can be semantically equivalent and thereby represent the same function, which introduces a bias. 
This problem has been addressed by \citet{biggio2021neural} and \citet{lalande23_transformer} who employ symbolic equivalence testing via \texttt{SymPy}~\citep{sympy_meurer17} and retain only simplified, canonical formulas for pre-training. 
Second, random sampling may fail to include functions that, despite having succinct representations, are not captured by the sampling process.

\textbf{Expression DAG enumeration.} More recent work has explored systematic enumeration of expression DAGs (directed acyclic graphs) to address the limitations of randomly sampling expression trees~\citep{kahlmeyer2024udfs}.
Expression DAGs can be used to compress expression trees by common subexpression detection and elimination. 
Therefore, expression DAGs provide more succinct representations than expression trees.
Systematically enumerating and retaining only semantically different DAGs covers more semantic equivalence classes of functions compared to an equivalent number of random trees.
This results in a more comprehensive and less biased pre-training dataset. 

\subsection{Sampling Strategy}
To create training instances, data points are sampled from the symbolic formulas. Here, a sampling strategy comprises the choice of sampling domain and the sampling distribution on this domain.

\textbf{Sampling domain.} \quad
Although the definition domains of many of the symbolic pre-training formulas are unbounded, the sampling domain 
will be bounded since the samples are finite. 
While \citet{biggio2021neural} and  \citet{vastl2024symformer} sample from a hypercube, \citet{kamienny22_transformer} sample from a domain that in principle is unbounded by sampling from a standard normal distribution.

\textbf{Sampling distribution.} \quad
\citet{vastl2024symformer} and \citet{lalande23_transformer} use different variants of uniform sampling. 
\citet{biggio2021neural} employ a two-stage sampling process: First, they uniformly sample a small hypercube within the hypercube that constitutes the pre-training domain; second, on the small hypercube, they uniformly sample the actual pre-training sample points. 
In order to maximize the diversity of input distributions seen during pre-training, \citet{kamienny22_transformer} have introduced a diverse sampling strategy that samples the pre-training data points from a mixture of distributions within the pre-training domain.
Moreover, to avoid overfitting to the pre-training domain, \citet{kamienny22_transformer} standardize the pre-training data points before they are fed into the transformer.
Since data in applications of symbolic regression can be noisy, noise has also been added to the pre-training data.
\citet{biggio2021neural}, \citet{kamienny22_transformer}, and \citet{vastl2024symformer} add Gaussian noise sampled from $\mathcal{N}(0, \epsilon)$ for a range of small $\epsilon$-values, typically $\varepsilon$ is at most $0.1$. 

\subsection{Evaluations and Benchmarks}
%

%
\textbf{Evaluating generalization abilities.} Most evaluations of pre-trained symbolic regression algorithms report accuracy and complexity measures on SRBench~\citep{srbench_lacava21}. Often high-accuracy solutions are counted as successful recovery.
However, SRBench’s notion of accuracy can be misleading, when the training (and test) data are drawn from the same domain as the pre-training data. 
A telling indicator of a potential generalization issue is the metric of ground-truth formula recovery. 
On SRBench, leading search-based methods can recover up to $50\%$~\citep{pysr_cranmer23, kahlmeyer2024udfs} of the ground-truth formulas, whereas a state-of-the-art transformer-based model recovers only $1.59\%$ \citep{kamienny22_transformer}. 
This stark difference suggests that current approaches may excel at interpolating within their pre-training domain but struggle to generalize to novel instances. To our knowledge, no work has systematically investigated the generalization capabilities of these approaches, a critical aspect for their application to real-world scientific discovery. Our work fills this gap by providing the first focused empirical analysis of this phenomenon.


\section{Methodology}
\label{sec:method}
\begin{figure}[t!]
    \centering
    \input{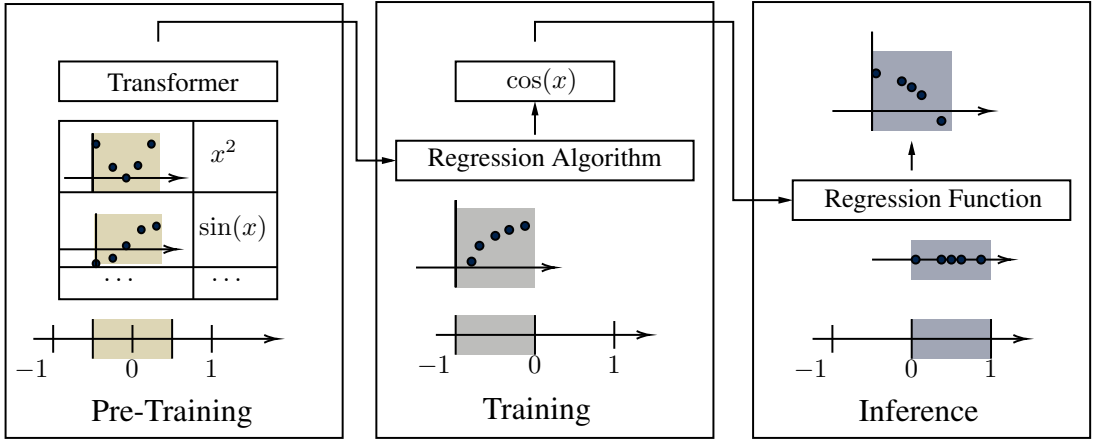}
    \caption{\textbf{Transformer-based symbolic regression.} Approaches turn pre-training data, consisting of symbolic formulas $f^i$ and samples $\big(x^{ij},y^{ij}=f(x^{ij})\big)$ from these formulas, into a regression algorithm. A regression algorithm maps training data $\big(x^k,y^k\big)$ into a regression function $f$, and a regression function $f$ maps feature vectors $x\in\mathbb{R}^d$ into labels $y=f(x)\in\mathbb{R}$.}
    \label{fig:taxonomy}
    \vspace{-1em}
\end{figure}

As illustrated in Figure~\ref{fig:taxonomy}, a transformer-based symbolic \emph{regression approach} turns \emph{pre-training data} into a regression algorithm, namely, a \emph{transformer model} with fixed parameters. \emph{Regression algorithms} map \emph{training data} into regression functions. A \emph{regression function} is a function that maps variables $x\in\mathbb{R}^n$ to labels $y\in\mathbb{R}$. That is, a transformer with fixed parameters is a regression algorithm that maps training data into regression functions.

In contrast to, e.g., search-based symbolic regression algorithms, transformer-based algorithms make use of a pre-training phase.
The \textbf{pre-training distribution} of a pre-trained approach is comprised by three components: (1) the set of \textbf{symbolic expressions}, (2) their \textbf{sampling domain} as well as (3) the \textbf{sampling distribution} with which they have been sampled. 

\paragraph{Symbolic Expressions.}
Evaluating the true generalization capabilities of pre-trained models requires testing on a held-out set of symbolic expressions, distinct from those seen during pre-training. A significant methodological challenge, however, is that the exact composition of the pre-training datasets for the four investigated state-of-the-art models is not publicly disclosed. To enable a controlled investigation, we train our own model from scratch. This approach grants us full control over the data-generation process, allowing us to create a well-defined test set of unseen expressions. This custom-trained model serves as the foundation for our ablation studies, a detailed analysis of which is presented in the supplementary material. The core of our empirical evaluation in the main paper, however, centers on the two other crucial axes of generalization: robustness to shifts in the sampling domain and distribution, which can be evaluated across all models.

\paragraph{Sampling domain.} The definition domain of many functions that symbolic regression seeks to recover is unbounded. Therefore, a finite number of samples used for pre-training will never cover the entire definition domain. In practice, the pre-training data will always only sample a bounded subset of the definition domain, because the number of samples is finite. Therefore, an assessment of the generalization performance of transformer-based symbolic regression algorithms must take their pre-training domain into account. 
To rigorously test generalization in practice, we create training data for each formula in SRBench on three distinct regimes: 

\begin{enumerate}
[
    label=\textbullet,
    topsep=0pt,         
    itemsep=4pt,        
    partopsep=0pt,      
    parsep=0pt,
    leftmargin=12pt
]
    \item \textbf{In-pre-training domain (IPD):} Training data is sampled from the same numerical domain used for the model's pre-training. For each model, the pre-training domain is given as a hypercube $[a,\, b]^d$, where $d$ is the number of variables. See supplementary material for model specific details. 
    \item \textbf{Within-pre-training-domain (WIPD):} Training data is sampled from a smaller hypercube fully contained within the pre-training domain. The hypercube is sampled as follows: First, a point is sampled uniformly at random from the pre-training domain. Then a side length $l$ is sampled uniformly from $[a,b]$, where $a$ and $b$ depend on the pre-training domain. The sampled hypercube is rejected when it is not completely contained in the pre-training domain. 
    \item \textbf{Out-of-pre-training-domain (OOPD):} Training data is sampled from a hypercube that is explicitly disjoint from the pre-training domain. This directly tests the model's extrapolation and generalization capabilities. The hypercube is sampled as follows: First, a point is sampled uniformly at random from the hypercube $[-100,100]^d$. Then a side length $l$ is sampled uniformly from $[a,b]$, where $a=1$ and $b=200$. The sampled hypercube is rejected when it overlaps with the pre-training domain.
\end{enumerate} 

To evaluate the accuracy performance of a regression function returned by an SR method, we follow standard machine learning practice:

\begin{enumerate}
[
    label=\textbullet,
    topsep=0pt,         
    itemsep=4pt,        
    partopsep=0pt,      
    parsep=0pt,
    leftmargin=12pt
]
    \item \textbf{In-Domain (ID):}  We test the regression function on data points drawn from the same domain as the training data points used to fit it (e.g., fit on $x \in [-5, 5]$ and test on new points $x \in [-5, -5]$).
    \item \textbf{Out-of-Domain (OOD):} We test the regression function on data from a different domain than the training data (e.g., fit on $x \in [-5, 5]$ but test on $x \in [6, 10]$).
\end{enumerate} 

\paragraph{Sampling Distribution.}
The distribution used to sample points from each symbolic expression is a critical design choice in the pre-training phase. To analyze its impact on model generalization, we evaluate performance under a controlled, parametric shift away from each model's native pre-training distribution. This is achieved by transforming an initial data point $x_{\text{init}}$ drawn from the original domain $[a, b]$ using the following power-law skew:
\begin{equation}
    x = a + (b-a) \left( \frac{x_{\text{init}} - a}{b-a} \right)^{\exp(-\epsilon)}
\end{equation}
The parameter $\epsilon$ controls the skew: $\epsilon=0$ recovers the original distribution, while $\epsilon > 0$ concentrates points towards the upper bound $b$, and $\epsilon < 0$ concentrates them towards the lower bound $a$.


\section{Experiments}
\label{sec:experiments}
Our empirical evaluation is designed to address two central research questions. First, we assess to what extent pre-trained symbolic regression models generalize when tested beyond the scope of their pre-training domain. Second, we investigate how robust these models are to common data perturbations encountered in practice such as domain shifts, distribution changes, and noise. 


\subsection{Experimental Setup}
\label{sec:experimental_setup}

\textbf{Models and baselines.} 
We evaluate the in-pre-training domain (IPD), within-pre-training domain (WIPD), and out-of-pre-training domain (OOPD) performance of five
pre-trained symbolic regression methods. Four are prominent, pre-trained, transformer models for which the pre-training domain is known: 
\textbf{NeSymReS}~\citep{biggio2021neural}, \textbf{E2E}~\citep{kamienny22_transformer}, \textbf{TF4SR}~\citep{lalande23_transformer}, and \textbf{SymFormer}~\citep{vastl2024symformer}. 
The fifth is \textbf{TPSR}~\citep{Shojaee2023TransformerbasedPF}, a hybrid approach that combines a pre-trained transformer with a subsequent Monte Carlo Tree Search (MCTS).
We benchmark these pre-trained methods against \textbf{PySR}~\citep{pysr_cranmer23}, a state-of-the-art, non-pre-trained search-based baseline. For a broader context, the supplementary material provides a comprehensive comparison against a larger set of recent symbolic regression algorithms.

\textbf{Datasets and metrics.}
Our evaluation is performed on the $144$ ground-truth formulas from the SRBench benchmark~\citep{srbench_lacava21}. For each problem in the benchmark, we train the given method using training data drawn from IPD, WIPD and OOPD. For each formula and regime, we sample 255 training data points using a uniform sampling strategy. We assess performance using two primary metrics. The first is \textbf{recovery rate}, which measures the percentage of SRBench formulas for which a method returns a symbolically equivalent expression to the ground truth. Following~\citet{srbench_lacava21}, we consider a ground truth formula $f$ recovered by an expression $g$ if their difference ($f-g$) or, for non-zero $g$, their ratio ($f/g$) can be symbolically resolved to a constant. All symbolic computations are performed using \texttt{SymPy}~\citep{sympy_meurer17}. The second metric is \textbf{accuracy}, which we quantify using the coefficient of determination ($R^2$) on the generated symbolic expressions. The performance is evaluated on two separate test sets: an ID set, whose samples are drawn from the same domain as the training data, and an OOD set from a disjoint domain.

\begin{figure}[t!] 
    \centering 

    \begin{subfigure}[t]{0.329\textwidth} 
        \centering
        \includegraphics[width=\linewidth]{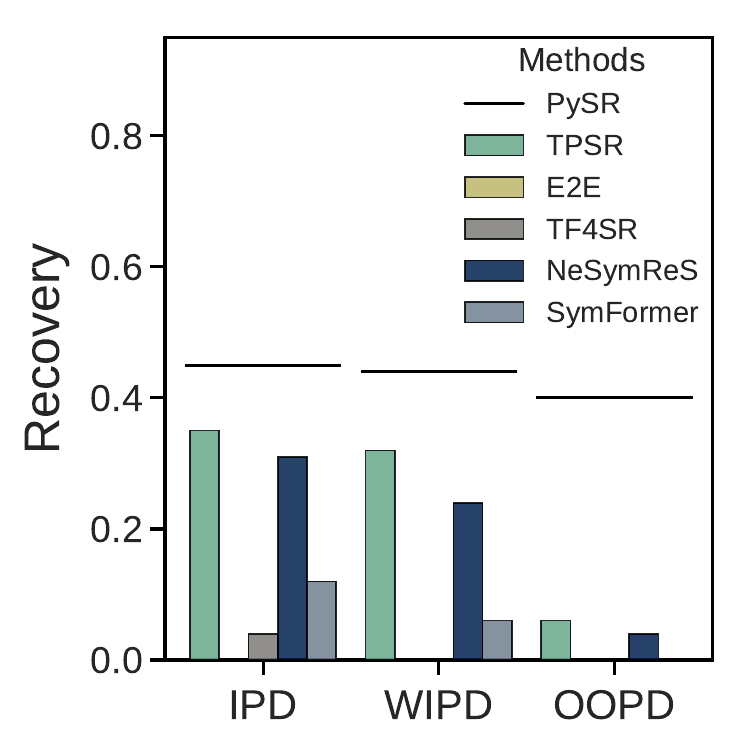} 
        \caption{Recovery}
        \label{fig:srbench_method_comparison}
    \end{subfigure}
    \hfill 
    \begin{subfigure}[t]{0.329\textwidth} 
        \centering
        \includegraphics[width=\linewidth]{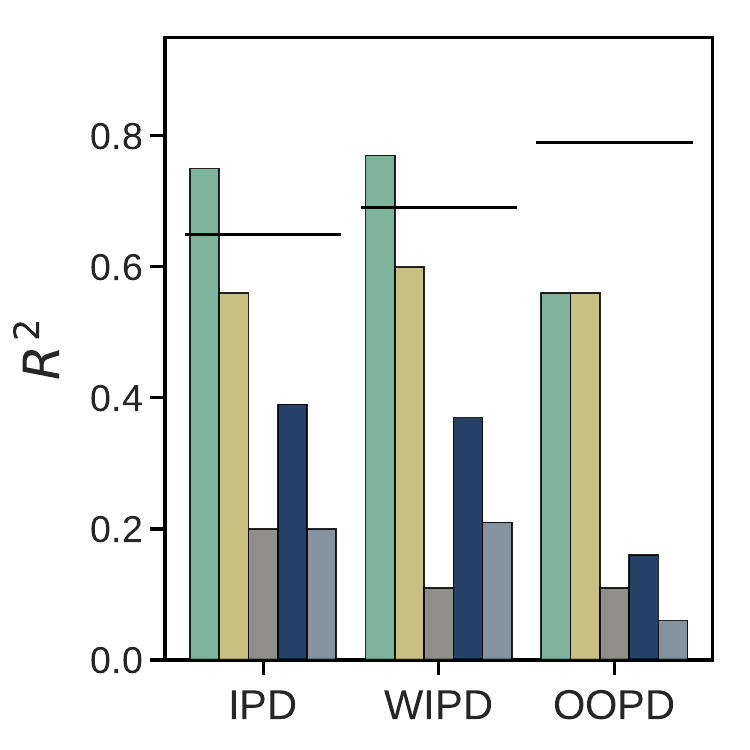} 
        \caption{Accuracy ID}
        \label{fig:accuracy_comparison_id}
    \end{subfigure}
    \hfill 
    \begin{subfigure}[t]{0.329\textwidth} 
        \centering
        \includegraphics[width=\linewidth]{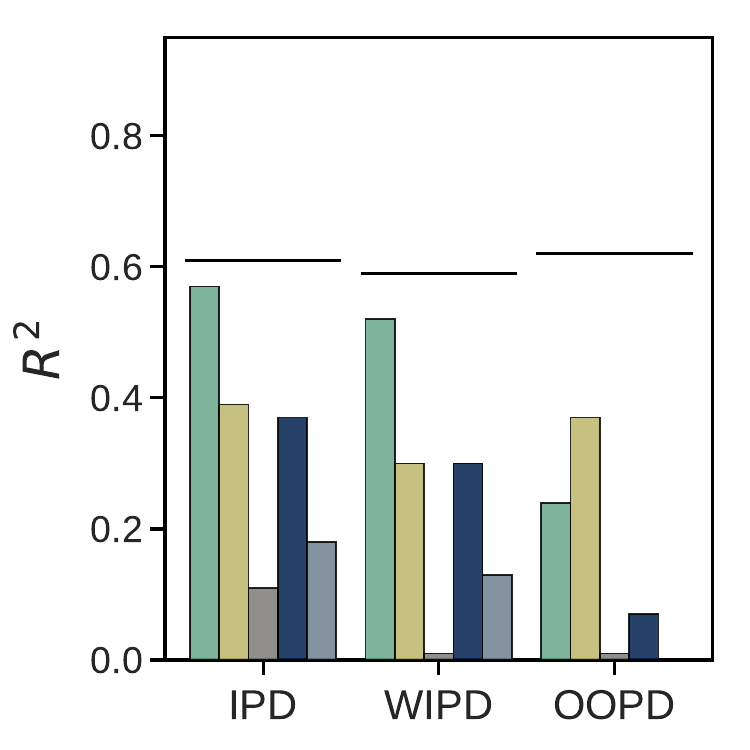} 
        \caption{Accuracy OOD}
        \label{fig:accuracy_comparison_od}
    \end{subfigure}
    \caption{\textbf{Recovery and accuracy performance on SRBench}. IPD, WIPD, and OOPD performance of TPSR, E2E, TF4SR, NeSymReS,  and SymFormer (transformer-based), and PySR (search-based baseline) on SRBench. The accuracy performance is measured on data points that have been sampled from the same domain as the training data (ID) and from a different domain (OOD).} 
    \label{fig:recovery_and_accuracy_comparison_on_srbench}
    \vspace{-1em}
\end{figure}

\subsection{Analysis of Out-of-pre-training-domain Generalization}
\label{sec:generalization_experiments}
We first evaluate the performance of the transformer-based approaches across the IPD, WIPD, and OOPD regimes. The results, summarized in  Figure~\ref{fig:srbench_method_comparison}, reveal a notable gap in generalization.

\textbf{IPD, WIPD, and OOPD recovery performance.}\quad 
All five transformer-based methods show significantly lower recovery performance than the search-based baseline PySR, which achieves a recovery rate of slightly more than $45\%$. 
Since PySR is not pre-trained its recovery rate does not depend on the training domains, IPD, WIPD, or OOPD. It remains fairly consistent across all three regimes. This does not hold for the transformer models. Among the pre-trained methods, NeSymReS has an IPD recovery rate of $31\%$ and SymFormer a recovery rate of $12\%$. The IPD recovery rates of TF4SR and E2E are close to zero. However, the recovery rates of NeSymReS and SymFormer markedly decline when transitioning from IPD to WIPD training data, and degrade further on OOPD training data. The same declining recovery performance is observed for the hybrid TPSR approach. 


\textbf{IPD, WIPD, and OOPD accuracy performance.}\quad 
The results of the ID accuracy experiments are summarized in Figure~\ref{fig:accuracy_comparison_id}. 
The search-based PySR baseline demonstrates consistent, high-accuracy performance across domains. E2E shows the best accuracy among the pre-trained transformer models. 
Its accuracy performance does not degrade when the training data move out of the pre-training domain. NeSymReS shows reasonable but signifcantly worse accuracy performance.
Its accuracy depends on the pre-training domain and degrades when the training data move out of the pre-training domain. The accuracy of TF4SR and SymFormer on all training data domains is low. 
The hybrid TPSR approach demonstrates the strongest accuracy performance in the IPD and WIPD regimes, but deteriorates in the OOPD regime. 
The results of the OOD accuracy experiments are summarized in Figure~\ref{fig:accuracy_comparison_od}. The search-based PySR baseline shows consistent accuracy performance in all three regimes. 
All transformer-based approaches show a substantial drop in accuracy on OOD test data, compared to the ID scenario. The E2E approach exhibits a lower, slightly fluctuating accuracy of around $0.4$. The other three pre-trained transformers show declining accuracy when moving from IPD to WIPD to OOPD. The same declining performance is observed for the hybrid TPSR approach.

\subsection{Case Study: Robustness to Data Perturbations}
\label{sec:case_study}
Our case study is designed to analyze model sensitivity to common data variations in the \textbf{training data} on the single, simple problem $y = x^2$. We systematically apply four types of perturbations: (1) scaling the input domain, (2) shifting the input domain, (3) adding Gaussian noise to the output values, and (4) varying the sampling distribution itself. This allows for a granular analysis of how these factors individually impact model performance. For details, see the \textbf{supplementary material}.

\begin{figure}[t!] 
    \centering 

    \begin{subfigure}[t]{0.49\textwidth} 
        \centering
        \includegraphics[height=4.5cm]{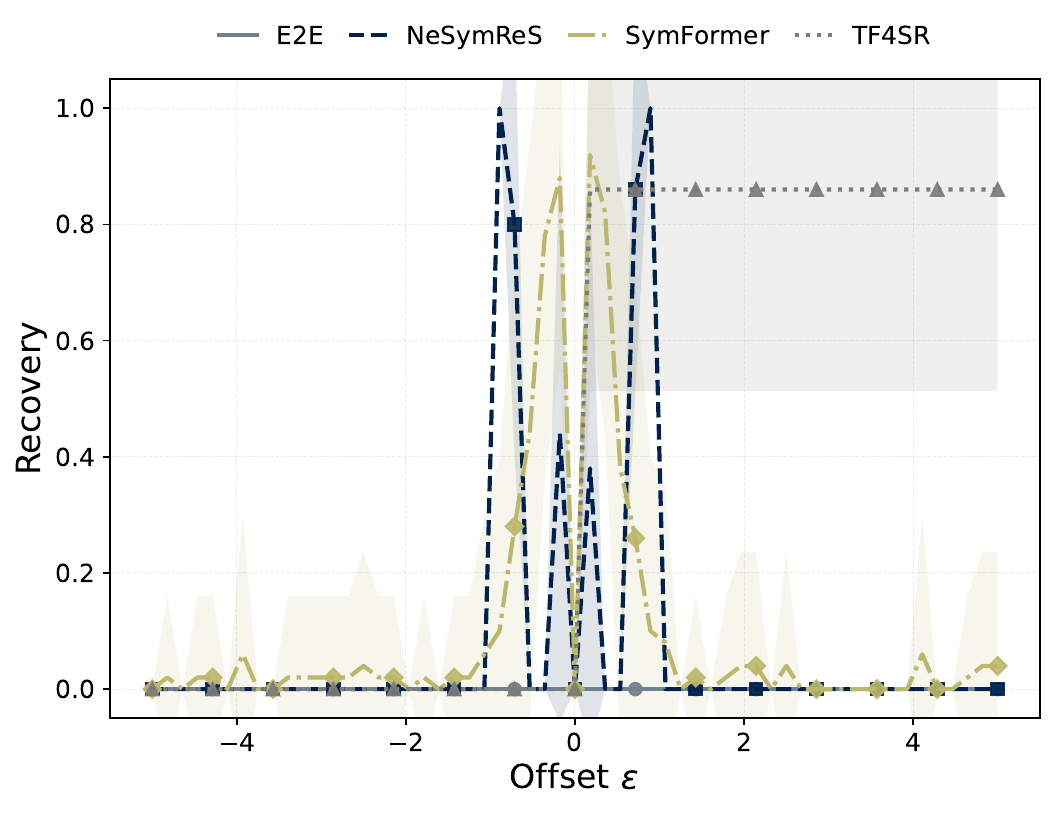}
        \caption{Input Scaling}
        \label{fig:robustness_transformers_scale_input}
    \end{subfigure}
    \hfill 
    \begin{subfigure}[t]{0.49\textwidth} 
        \centering
        \includegraphics[height=4.5cm]{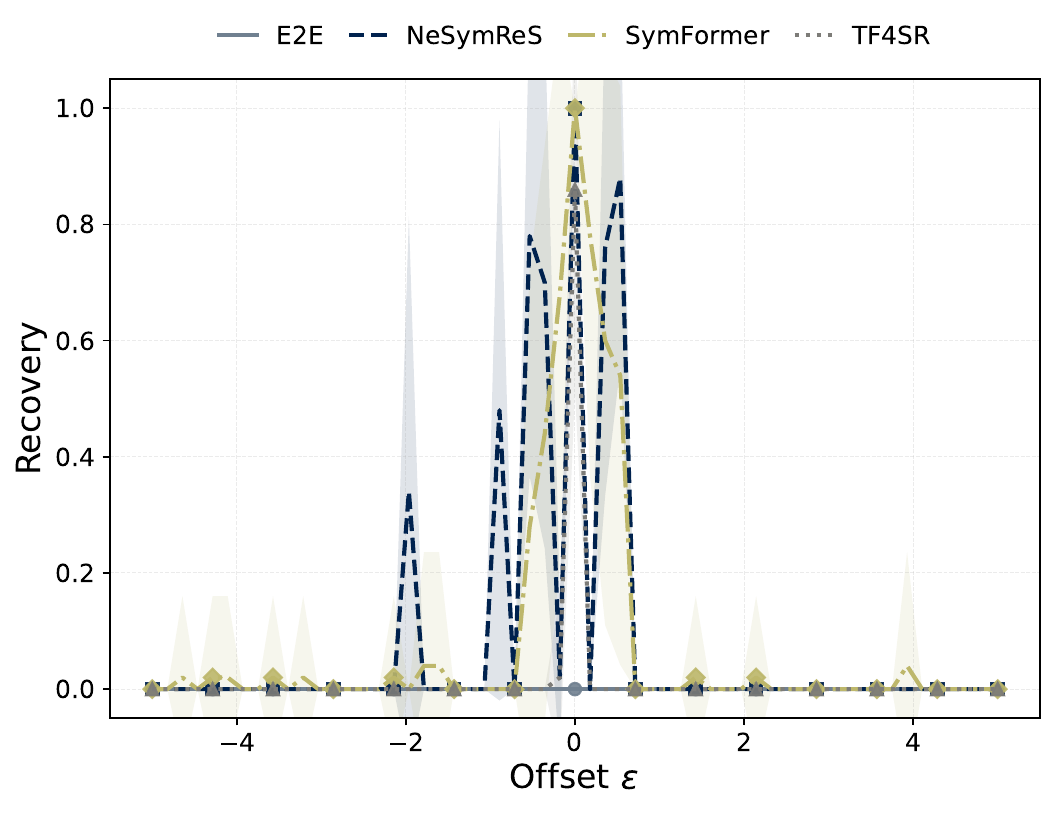}
         \caption{Input Shift}
        \label{fig:robustness_transformers_shift_input}
    \end{subfigure}
    \begin{subfigure}[t]{0.49\textwidth} 
        \centering
        \includegraphics[height=4.5cm]{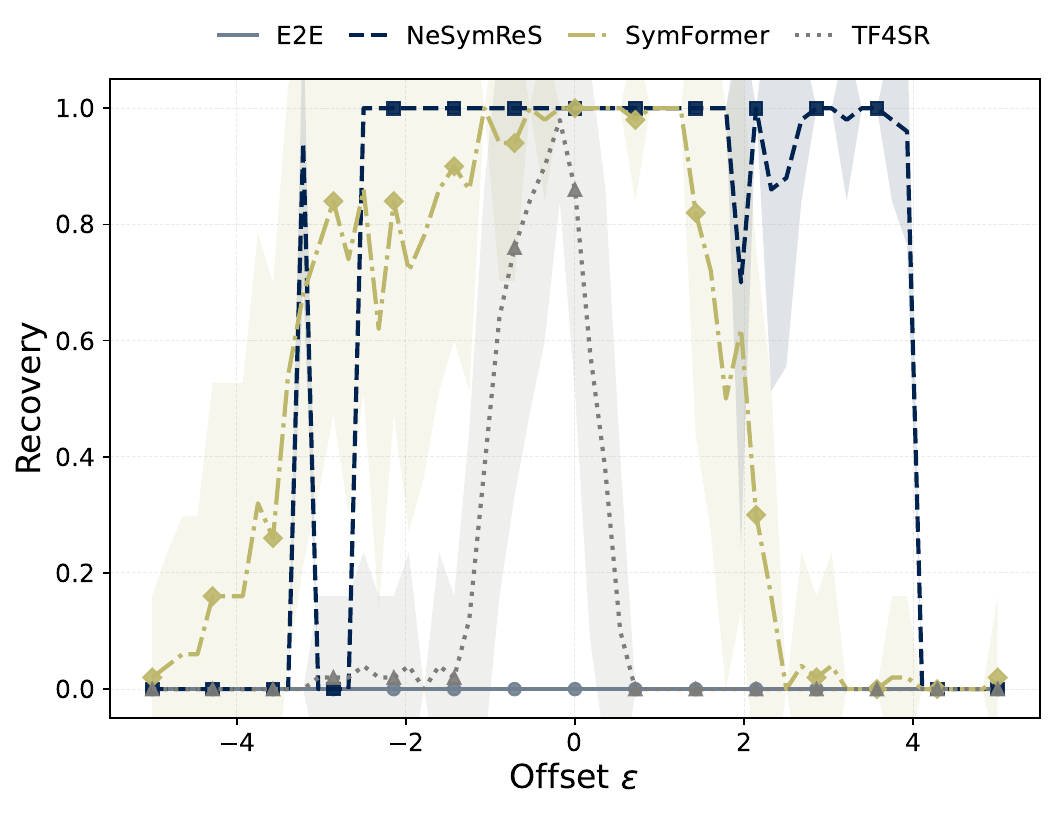}
        \caption{Distribution Shift}
        \label{fig:robustness_transformers_distribution}
    \end{subfigure}
    \hfill 
    \begin{subfigure}[t]{0.49\textwidth} 
        \centering
        \includegraphics[height=4.5cm]{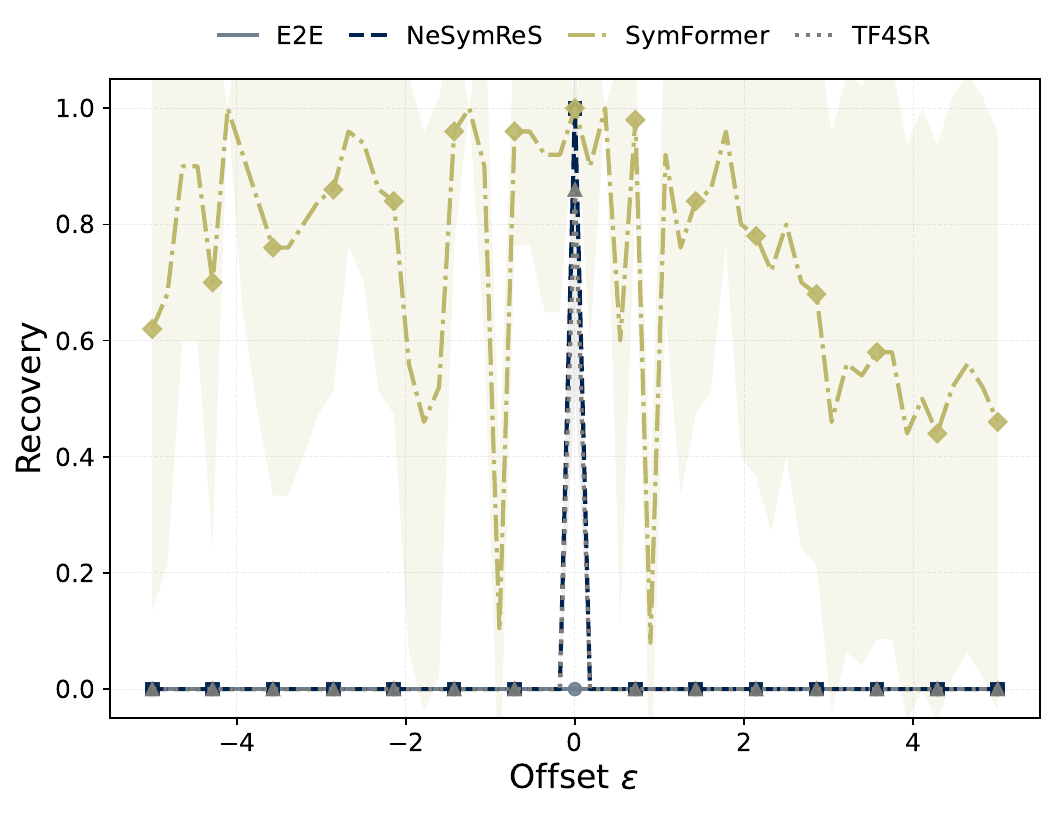}
        \caption{Output Noise}
        \label{fig:robustness_transformers_noise}
    \end{subfigure}
    \caption{
        \textbf{Robustness analysis reveals widespread fragility.}
        The figure shows the recovery rate of four pre-trained transformer models on the simple problem $y=x^2$ under four types of data perturbations. The parameter $\epsilon$ controls the magnitude of the perturbation, with $\epsilon=0$ being the unperturbed case. Lines represent the mean recovery over $50$ trials, with shaded areas indicating the standard deviation. A primary finding is the extreme brittleness of all models. While some models like NeSymReS and SymFormer show broader robustness to distribution shifts, they remain highly sensitive to input scaling and input shifts.
    }
    \label{fig:transformer_robustness_comparison}
\end{figure}

\noindent The case study results, presented in Figure~\ref{fig:transformer_robustness_comparison}, reveal a widespread and critical lack of robustness across all tested models, with performance being highly sensitive to the type of perturbation.

For perturbations to the \textbf{input domain} (Fig.~\ref{fig:transformer_robustness_comparison}a, b), successful recovery is confined to an extremely narrow window around the unperturbed case ($\epsilon=0$). For both input scaling and shifting, NeSymReS and SymFormer exhibit sharp performance peaks that collapse almost immediately. TF4SR is a notable exception under scaling, maintaining high recovery for positive $\epsilon$ values, but it remains brittle to shifts. The models are most vulnerable to \textbf{output noise} (Fig.~\ref{fig:transformer_robustness_comparison}d), where even minimal noise leads to a catastrophic failure for nearly all models; only SymFormer shows a semblance of resilience, though its performance is erratic.

In contrast, some models demonstrate broader robustness to \textbf{distribution shifts} (Fig.~\ref{fig:transformer_robustness_comparison}c). NeSymReS and SymFormer, in particular, maintain high recovery across a wide range of negative $\epsilon$ values, while SymFormer's performance degrades asymmetrically. Overall, these findings highlight that even on a simple problem, pre-trained models are exceptionally brittle. Their success is contingent on the training data being almost perfectly aligned with their pre-training distributions, posing a significant challenge for real-world deployment.

\section{Discussion}
\label{sec:discussion}
Our empirical investigations reveal a gap of current pre-trained approaches to generalize to problems outside their pre-training distributions. The following discusses our main findings. 

\subsection{Memorizing Patterns vs. Learning Compositional Building Blocks}

Our results strongly suggest that current transformer-based approaches do not learn compositional, generalizable representations of the underlying data. Instead, they function as highly effective pattern matchers. During pre-training, they appear to learn a mapping between the statistical properties of numerical point clouds and the symbolic structures that generated them. During inference, they succeed only when a new problem's point cloud closely matches a pattern in their memorized library of structural point cloud snapshots. This explains the observed dichotomy in performance.
The high in-domain accuracy of models like E2E, which mirrors conventional regressors like polynomial regression, demonstrates their power as flexible interpolators. This behaviour can be explained by the architecture of E2E, which predicts a large number of free coefficients and is therefore highly flexible in fitting the data.  
This high accuracy, however, is misleading as it is achieved without identifying the true symbolic structure. 
The stark collapse in formula recovery when moving from IPD to OOPD problems is the clearest evidence of this generalization gap. 
Additionally, the learned representations prove brittle under simple perturbations, a phenomenon almost certainly present in any real-world data.  




\subsection{Why Technical Fixes Like Standardization Are Not a Solution}
One might argue that techniques like input standardization could alleviate this domain dependency. However, our results show this is not the case. 
The E2E approach, which relies on data standardization, exhibits a clear trade-off: it achieves strong IPD accuracy but performs poorly on symbolic recovery and OOPD accuracy. 
We explain the observed behavior as follows: Standardization simply transforms the numerical domain; it does not alter the underlying structural relationship between the data and the formula. Any pre-training method, by sampling points, takes a structural snapshot of a function in some pre-training domain. Even if the data is standardized and the model is trained on standardized data, then this standardized data still reflects the same structural snapshot of the underlying function taken from the pre-training domain. 
Consequently, models trained on standardized data recover expressions that behave identically to the snapshots the model has seen during training. However, for expressions that alter their behavior outside of the training data, the models lose their ability to recover.
What we observe is that the model mereley memorizes a structural snapshot from this pre-training domain, and standardization does not help it reason about structures it has never seen. The problem is not one of numerical range but of compositional generalization. 

\subsection{Implications for Practitioners}
%
Our findings provide a clear decision-making framework for the applied use of transformer-based symbolic regression methods: 

\begin{enumerate}
[
    label=\textbullet,
    topsep=0pt,         
    itemsep=4pt,        
    partopsep=0pt,      
    parsep=0pt,
    leftmargin=12pt
]
    \item For problems where the data is expected to be \textbf{well-contained within a model's pre-training domain}, a transformer-based method can be a remarkably fast tool for generating initial hypotheses.

    \item For \textbf{novel problem instances or data that deviates from the pre-training distribution}, our results serve as a strong cautionary note. Relying on these models risks generating misleading formulas that appear accurate locally but are not generalizing well. In these common real-world scenarios, robust search-based methods remain the more reliable choice.
\end{enumerate}

\subsection{Future Directions for Research}

This work illuminates a clear and necessary path for future investigations:

\begin{enumerate}
[
    label=\textbullet,
    topsep=0pt,         
    itemsep=4pt,        
    partopsep=0pt,      
    parsep=0pt,
    leftmargin=12pt
]
    \item \textbf{Data scaling is not the answer:} The results of our ablation study indicate that simply creating larger or more diverse pre-training datasets is unlikely to solve the fundamental generalization problem. This suggests that innovations in modelling are needed.

    \item \textbf{Focusing on OOPD robustness:} 
    Current symbolic regression approaches are mainly evaluated using in-domain metrics. A future direction is stronger prioritization of the development of models that can pass rigorous OOPD generalization tests. This involves evaluating pre-trained models in and out of their pre-training domains (IPD vs. OOPD), as well as rigorously evaluating the generated regression functions in and out of their training domains (ID vs. OOD).

    \item \textbf{Combining search and pre-trained approaches:} A promising research direction is the development of hybrid approaches that combine the robustness of search-based methods with the fast, informative prior provided by transformer-based approaches. A critical, open problem that needs to be solved here is ensuring that the learned prior is robust enough to reliably guide the search process, particularly in OOPD scenarios. Current hybrid methods such as TPSR represent a first step in this direction, despite the fact that, as our experiments have shown, the generated priors still suffer from generalization issues. More research is required to ensure that these fast priors provide reliable guidance in OOPD cases.
\end{enumerate}

\Forward\, In conclusion, while transformers have brought impressive scaling to symbolic regression, their practical value is currently capped by their pre-training data distribution. Overcoming this barrier is the central challenge for the next generation of pre-trained symbolic regression.

\section{Limitations}
\label{sec:limitations}
Our study is conducted on the widely used SRBench benchmark. While valuable for standardized comparison, SRBench is primarily composed of small-to-medium complexity formulas drawn from physics. The domain specificity means that generalization to other scientific fields with different characteristic mathematical structures (e.g., economics, chemistry) is not explicitly tested. Further, we evaluated five prominent and representative transformer-based models. While these cover key approaches in the field, the rapid pace of research means our findings may not extend to all possible future architectures. Despite these scope limitations, our central conclusion—the existence of a severe generalization gap when models are applied outside their pre-training distribution is a consistent and robust finding across all tested models and configurations. This highlights a fundamental challenge for the current generation of pre-trained symbolic regression, underscoring the need for both more diverse benchmarks and novel modeling approaches designed explicitly with generalization in mind.


\section{Conclusions}
\label{sec:conclusions}
Transformer-based models have introduced a powerful, scalable paradigm to symbolic regression, amortizing the high cost of combinatorial search through large-scale pre-training. 
In this work, we provided the first systematic evidence that this paradigm suffers from a critical limitation: a severe gap to generalize to problems outside of the pre-training distribution. 
Our results suggest that current models learn to recognize statistical patterns within their pre-training distribution rather than learning underlying compositional building blocks. 
This severely limits their reliability for real-world scientific discovery, where data is almost always observed across diverse domains and scales, and in noisy conditions.
Future research must focus on increasing the generalization ability of these methods.
A promising path forward then lies in combining the structured exploration of search-based algorithms with the fast inference times of transformers. 
Forging this synthesis is a crucial next step toward creating truly generalizing symbolic regression methods to support scientific discovery.

%
%
%



\bibliography{iclr2026_conference}
\bibliographystyle{iclr2026_conference}



\clearpage


\section*{Supplementary material}

# Analyzing Generalization in Pre-Trained Symbolic Regression

# Supplementary Material

## Contents

# 1 Generalization Experiments

This section provides a detailed analysis of our generalization experiments, organized as follows. First, we precisely define the pre-training domain for each method under investigation. Second, we present a comprehensive results table, benchmarking the pre-trained and hybrid models against a suite of baselines, including search-based methods and classical regressors. Finally, we conduct an analysis of the performance on specific subsets of the benchmark to characterize distinct behaviors and failure modes.

## 1.1 Experimental Setup

In our generalization experiments, we examine five well-known, publicly available, pre-trained, transformer-based symbolic regression approaches, for which the domains of pre-training are known.

**Pre-training data.** All methods employ the random tree generation procedure outlined in [Lample and Charton, 2019] to produce a large collection of synthetic training expressions. The pre-training domain of NeSymReS is the hypercube $[-10, 10]^d$, where $d$ is the number of variables. The pre-training domain of TF4SR is the hypercube $[0.1, 10]^d$. The pre-training domain of E2E is in principle unbounded, but more than 99% of the pre-training sample points are contained in the hypercube $[-3, 3]^d$. SymFormer is pre-trained within $[-5, 5]^d$. The evaluated methods employ distinct data sampling strategies that vary in complexity. SymFormer uses the most straightforward approach, drawing samples uniformly at random from within its hypercube. In a similar vein, TF4SR samples from the entire hypercube but employs a non-uniform logarithmic distribution, which biases samples toward the origin. NeSymReS adopts a structured two-stage procedure, first selecting a random subcube and then sampling uniformly from within it. Finally, E2E implements the most flexible strategy by constructing a mixture of base distributions (e.g., Gaussian and uniform), an approach capable in principle of approximating any target distribution. The hybrid TPSR approach utilises the pre-trained E2E transformer model as an informative prior, thereby being constrained to the pre-training data, domain and distribution of this approach.

**Varying input dimensions.** SymFormer has been pre-trained only on formulas with up to two variables ($d = 2$), NeSymReS on formulas with up to three variables ($d = 3$), TF4SR on formulas with up to six variables ($d = 6$), and E2E on formulas with up to ten variables ($d = 10$). In our experiments, we measure the recovery and the accuracy performance on the 144 ground truth formulas from SRBench [La Cava et al., 2021]. A key challenge in this benchmark is a dimensionality mismatch: some formulas contain more input variables than certain transformer models were pre-trained on. To ensure a fair comparison across all methods, we therefore stratify our evaluation by the number of input variables. Consequently, we report results on the full dataset (containing formulas with up to 10 dimensions) as well as on two lower-dimensional subsets, restricted to formulas with exactly 2 and 3 dimensions, respectively.

**Metrics.** Beyond the primary metrics of symbolic recovery and accuracy ($R^2$), we introduce additional metrics to provide a more nuanced analysis of symbolic structure and solution complexity. To supplement the strict, binary nature of the recovery metric, we report two measures of symbolic similarity for cases where an exact match is not found. The first is the **Tree Edit Distance (TED)**,

which quantifies the number of operations (e.g., insert, delete) required to transform the predicted expression tree into the ground truth. Here, we follow the implementation of Matsubara et al. [2024]. A lower TED indicates a closer structural match. The second is the **Jaccard Index (JI)**, which measures the similarity between the sets of all subtrees in the predicted and ground truth expressions, with higher values indicating greater overlap. We follow the implementation of the JI by Kahlmeyer et al. [2024]. Finally, to assess the parsimony and interpretability of the solutions, we report their **complexity**, defined as the number of operator nodes in the expression tree, following La Cava et al. [2021].

## 1.2 Results

Table 1 and Table 2 display the recovery and accuracy performance of the evaluated transformer-based approaches against the search-based and classical baselines.

### 1.2.1 Recovery Results

The results of the recovery experiments are summarized in Table 1. A primary finding is that all pre-trained transformer-based methods significantly underperform the leading search-based method, PySR, which achieves a recovery rate of over 45%. As a non-pre-trained method, PySR's performance is inherently robust to the IPD, WIPD, and OOPD distinctions. In contrast, the transformer models exhibit strong domain dependence. Among them, NeSymReS achieves the highest IPD recovery rate at 31%, whereas the rates for TF4SR and E2E are near zero. However, NeSymReS's performance declines markedly from IPD to WIPD and degrades further in the OOPD setting, highlighting its limited generalization. The evaluation on lower-dimensional subsets reveals that performance can improve when the problem dimensionality matches the model's strengths. For instance, on the 2-variable subset, SymFormer-2 achieves a $52\%$ recovery rate, competitive with NeSymReS-2. Nonetheless, a clear trend persists across all pre-trained models and subsets: recovery performance consistently deteriorates when moving from the pre-training domain to OOPD scenarios. Overall, NeSymReS demonstrates the strongest recovery performance among the transformer-based approaches. The hybrid TPSR method improves upon its E2E prior, achieving a $35\%$ IPD recovery rate where E2E alone scored zero. However, it inherits the same critical weakness, as its performance also declines sharply OOPD. The search-based methods showcase remarkable domain-independent stability. PySR is the top performer overall with a $45\%$ recovery rate. An exhaustive search provides a stable $35\%$ baseline, while the reinforcement learning-based DSR achieves a consistent $28\%$. Genetic algorithms like Operon ($23\%$) and GPLearn ($9\%$) achieve lower rates, but their performance is also stable across all domains. The stability of approaches like DSR suggests a promising direction for future research: online training or fine-tuning pre-trained models on a per-problem basis may be a crucial technique to overcome the OOPD generalization challenge. Finally, the classical baselines provide important context. Polynomial regression, a simple and computationally cheap method, achieves a $13\%$ recovery rate. This result underscores the challenge for complex pre-trained models. The substantial effort invested in their training should warrant performance that decisively surpasses such simple baselines. In summary, these comprehensive results indicate that significant work remains in improving the OOPD generalization and overall effectiveness of pre-trained symbolic regression approaches compared to their search-based and classical counterparts.

Table 1: **Symbolic recovery and ground truth similarity of all evaluated methods.** We compare pre-trained transformers, hybrid methods, and various search-based and classical baselines. Performance is evaluated across three domains: in-pre-training (IPD), within-pre-training (WIPD), and out-of-pre-training (OOPD). We report the exact **Recovery Rate** (higher is better), the normalized **Tree Edit Distance** to the ground truth (lower is better), and the **Jaccard Index** of the expression skeletons (higher is better). The suffixes **-3** and **-2** denote evaluations on subsets of the benchmark restricted to formulas with exactly 3 and 2 input variables, respectively. The results highlight the superior robustness of search-based methods like PySR, especially in OOPD scenarios.

|              | Recovery |      |      | Tree Edit Distance |      |      | Jaccard Index |      |      |
| ------------ | -------- | ---- | ---- | ------------------ | ---- | ---- | ------------- | ---- | ---- |
| **Method**   | IPD      | WIPD | OOPD | IPD                | WIPD | OOPD | IPD           | WIPD | OOPD |
| **Transformer** |       |      |      |                    |      |      |               |      |      |
| NeSymRes     | 0.31     | 0.24 | 0.04 | 0.73               | 0.75 | 0.92 | 0.25          | 0.25 | 0.12 |
| E2E          | 0.00     | 0.00 | 0.00 | 0.99               | 0.99 | 0.99 | 0.14          | 0.14 | 0.13 |
| TF4SR        | 0.04     | 0.00 | 0.00 | 0.73               | 0.84 | 0.82 | 0.29          | 0.22 | 0.22 |
| SymFormer    | 0.12     | 0.06 | 0.00 | 0.87               | 0.92 | 0.99 | 0.11          | 0.09 | 0.03 |
| NeSymRes-3   | 0.62     | 0.49 | 0.08 | 0.46               | 0.48 | 0.83 | 0.51          | 0.50 | 0.25 |
| E2E-3        | 0.00     | 0.00 | 0.00 | 0.98               | 0.98 | 0.98 | 0.13          | 0.13 | 0.11 |
| TF4SR-3      | 0.07     | 0.00 | 0.00 | 0.73               | 0.90 | 0.84 | 0.28          | 0.18 | 0.19 |
| SymFormer-3  | 0.24     | 0.11 | 0.00 | 0.75               | 0.85 | 0.98 | 0.23          | 0.18 | 0.06 |
| NeSymRes-2   | 0.64     | 0.55 | 0.18 | 0.45               | 0.48 | 0.76 | 0.53          | 0.51 | 0.28 |
| E2E-2        | 0.00     | 0.00 | 0.00 | 0.97               | 0.96 | 0.97 | 0.12          | 0.12 | 0.11 |
| TF4SR-2      | 0.06     | 0.00 | 0.00 | 0.72               | 0.87 | 0.86 | 0.25          | 0.17 | 0.15 |
| SymFormer-2  | 0.52     | 0.24 | 0.00 | 0.45               | 0.67 | 0.97 | 0.50          | 0.38 | 0.13 |
| **Hybrid**   |          |      |      |                    |      |      |               |      |      |
| TPSR         | 0.35     | 0.32 | 0.06 | 0.71               | 0.74 | 0.93 | 0.32          | 0.29 | 0.16 |
| **Search**   |          |      |      |                    |      |      |               |      |      |
| Exhaustive   | 0.35     | 0.35 | 0.35 | 0.53               | 0.54 | 0.54 | 0.48          | 0.48 | 0.49 |
| PySR         | 0.45     | 0.44 | 0.40 | 0.58               | 0.60 | 0.62 | 0.46          | 0.46 | 0.43 |
| Operon       | 0.23     | 0.22 | 0.17 | 0.80               | 0.82 | 0.85 | 0.21          | 0.21 | 0.19 |
| GPlearn      | 0.09     | 0.11 | 0.09 | 0.90               | 0.82 | 0.88 | 0.16          | 0.22 | 0.19 |
| DSR          | 0.28     | 0.28 | 0.26 | 0.73               | 0.71 | 0.75 | 0.36          | 0.36 | 0.34 |
| **Classical**|          |      |      |                    |      |      |               |      |      |
| LinReg       | 0.01     | 0.01 | 0.01 | 0.92               | 0.92 | 0.92 | 0.19          | 0.19 | 0.19 |
| PolyReg      | 0.13     | 0.13 | 0.13 | 0.90               | 0.90 | 0.89 | 0.14          | 0.13 | 0.15 |

#### 1.2.2  Accuracy Results

The results of the accuracy experiments are summarized in Table 2.

**ID accuracy results.** The results of the in-domain (ID) accuracy experiments are summarized in the left column of Table 2. Contrary to the recovery results, E2E shows the best accuracy among the transformer models. Moreover, accuracy performance does not degrade when the training data move

Table 2: **Predictive accuracy (R²) and solution complexity across all methods.** We report the coefficient of determination ($R^2$, where higher is better) and the complexity of the resulting symbolic expression. Performance is evaluated separately on an **in-domain (ID) test set** and an **out-of-domain (OOD) test set** for each of the three training data regimes: in-pre-training (IPD), within-pre-training (WIPD), and out-of-pre-training (OOPD). The suffixes **-3** and **-2** denote evaluations on subsets of the benchmark restricted to formulas with exactly 3 and 2 input variables, respectively. The results highlight a significant performance gap between ID and OOD tests for many pre-trained models, contrasting with the more stable performance of search-based methods like PySR.

|  | Accuracy-ID (R²) | | | Accuracy-OOD (R²) | | | Complexity | | |
|---|---|---|---|---|---|---|---|---|---|
| **Method** | IPD | WIPD | OOPD | IPD | WIPD | OOPD | IPD | WIPD | OOPD |
| **Transformer** | | | | | | | | | |
| NeSymRes | 0.39 | 0.37 | 0.16 | 0.37 | 0.30 | 0.07 | 4.21 | 4.54 | 6.39 |
| E2E | 0.56 | 0.60 | 0.56 | 0.39 | 0.30 | 0.37 | 20.38 | 20.64 | 20.82 |
| TF4SR | 0.20 | 0.11 | 0.11 | 0.11 | 0.01 | 0.01 | 6.56 | 7.60 | 6.91 |
| SymFormer | 0.20 | 0.21 | 0.06 | 0.18 | 0.13 | 0.00 | 5.07 | 7.03 | 9.25 |
| NeSymRes-3 | 0.80 | 0.76 | 0.32 | 0.75 | 0.62 | 0.14 | 4.21 | 4.54 | 6.39 |
| E2E-3 | 0.73 | 0.74 | 0.66 | 0.49 | 0.45 | 0.46 | 14.49 | 13.87 | 15.21 |
| TF4SR-3 | 0.31 | 0.15 | 0.15 | 0.17 | 0.01 | 0.00 | 5.18 | 7.88 | 7.00 |
| SymFormer-3 | 0.40 | 0.43 | 0.12 | 0.37 | 0.27 | 0.00 | 5.07 | 7.03 | 9.25 |
| NeSymRes-2 | 0.84 | 0.76 | 0.36 | 0.76 | 0.60 | 0.24 | 4.03 | 4.29 | 5.53 |
| E2E-2 | 0.72 | 0.81 | 0.77 | 0.55 | 0.50 | 0.60 | 12.58 | 10.45 | 12.61 |
| TF4SR-2 | 0.30 | 0.14 | 0.16 | 0.14 | 0.00 | 0.00 | 4.39 | 6.66 | 6.54 |
| SymFormer-2 | 0.87 | 0.92 | 0.26 | 0.80 | 0.58 | 0.00 | 5.07 | 7.03 | 9.25 |
| **Hybrid** | | | | | | | | | |
| TPSR | 0.75 | 0.77 | 0.56 | 0.57 | 0.52 | 0.24 | 13.60 | 13.62 | 16.47 |
| **Search** | | | | | | | | | |
| Exhaustive | 0.52 | 0.50 | 0.58 | 0.47 | 0.43 | 0.47 | 4.14 | 4.20 | 4.16 |
| PySR | 0.65 | 0.69 | 0.79 | 0.61 | 0.59 | 0.62 | 7.20 | 8.07 | 7.82 |
| Operon | 0.54 | 0.63 | 0.61 | 0.41 | 0.40 | 0.48 | 20.54 | 20.88 | 21.55 |
| GPlearn | 0.17 | 0.20 | 0.20 | 0.16 | 0.18 | 0.13 | 34.67 | 25.16 | 42.36 |
| DSR | 0.52 | 0.56 | 0.58 | 0.44 | 0.40 | 0.38 | 8.94 | 8.32 | 9.18 |
| **Classical** | | | | | | | | | |
| LinReg | 0.11 | 0.18 | 0.33 | 0.06 | 0.06 | 0.05 | 7.34 | 7.39 | 7.32 |
| PolyReg | 0.28 | 0.35 | 0.40 | 0.13 | 0.13 | 0.17 | 269.02 | 285.32 | 195.36 |

out of the pre-training domain. That is, as expected from a good regressor, E2E performs well in terms of accuracy on ID test data, independently from where the training data have been sampled. NeSymReS also shows reasonable but significantly worse accuracy performance. The accuracy of NeSymReS, however, depends on the pre-training domain. It degrades when the training data move out of the pre-training domain. Both recovered and missed formulas contribute to the overall accuracy, and the accuracy of recovered formulas is perfect by definition. Therefore, the fact that NeSymReS, the transformer with the best recovery, achieves only a moderate overall accuracy means that its accuracy

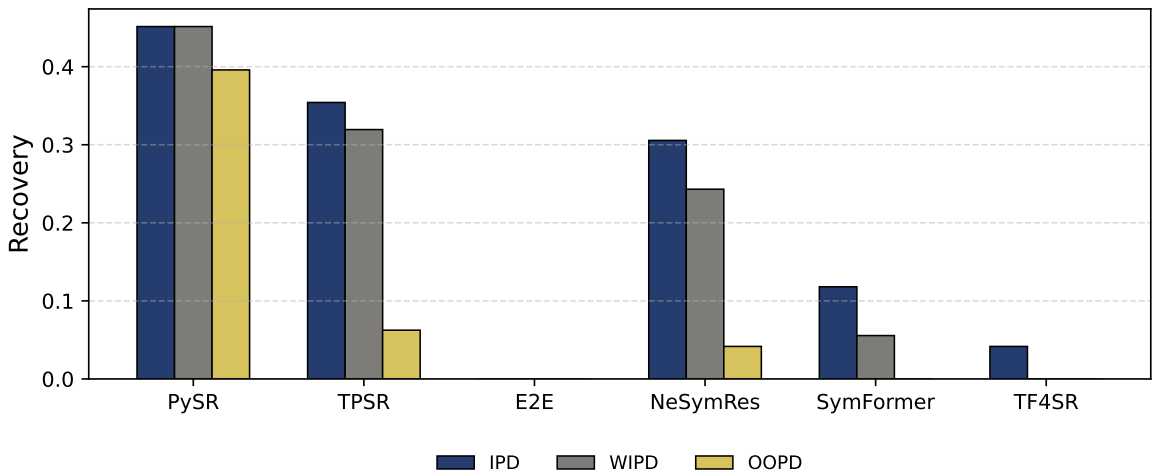

Figure 1: **Symbolic recovery performance across different domains.** The figure compares the recovery rate (higher is better) of the search-based baseline (PySR), the hybrid model (TPSR), and several pre-trained transformers. Performance is evaluated across three domains: in-pre-training (IPD), within-pre-training (WIPD), and out-of-pre-training (OOPD). A clear trend emerges: while pre-trained models like NeSymRes as well as hybrid methods like TPSR perform reasonably well IPD and WIPD, their recovery rate drops sharply in OOPD scenarios. In contrast, the search-based PySR maintains a high and stable recovery rate across all domains. This highlights the significant OOPD generalization challenge of pre-trained symbolic regression approaches.

on missed formulas will be markedly worse. The accuracy of TF4SR on all training data domains is low. The hybrid and search-based methods highlight a crucial trade-off between peak in-domain performance and robust generalization. The hybrid TPSR method demonstrates the potential of combining pre-trained models with search; leveraging its E2E prior, it achieves the highest ID accuracy of any method within its pre-training domain (IPD/WIPD). However, this performance is not robust, as its accuracy drops significantly in the OOPD case. This highlights a critical challenge for hybrid models: their success is contingent on the OOPD generalization of the underlying pre-trained component, which must be strong enough to reliably guide the search. In contrast, the pure search-based methods exhibit greater stability. PySR leads this category in ID accuracy, followed by strong, consistent performance from exhaustive search, Operon, and DSR. Their high accuracy is intrinsically linked to their recovery capabilities: successfully recovering a formula guarantees a perfect $R^2$ score for that instance, which increases their overall average performance across all domains.

**OOD accuracy results.** The OOD accuracy results, summarized in the central columns of Table 2, reveal a critical distinction between methods that find the true symbolic form and those that merely fit the data. A recovered formula naturally has perfect accuracy on any test set, making its performance independent of the data domain. This principle explains the stability of PySR: due to its high recovery rate, its accuracy shows minimal degradation when shifting from ID to OOD test data. In stark contrast, the pre-trained transformers behave like conventional interpolators. NeSymReS and E2E, despite achieving high accuracy on ID data, suffer a substantial performance drop on the OOD test set. This failure to extrapolate is characteristic of classical methods like polynomial regression, which also exhibits a similar decline, suggesting that the transformers are overfitting to the training domain.

6

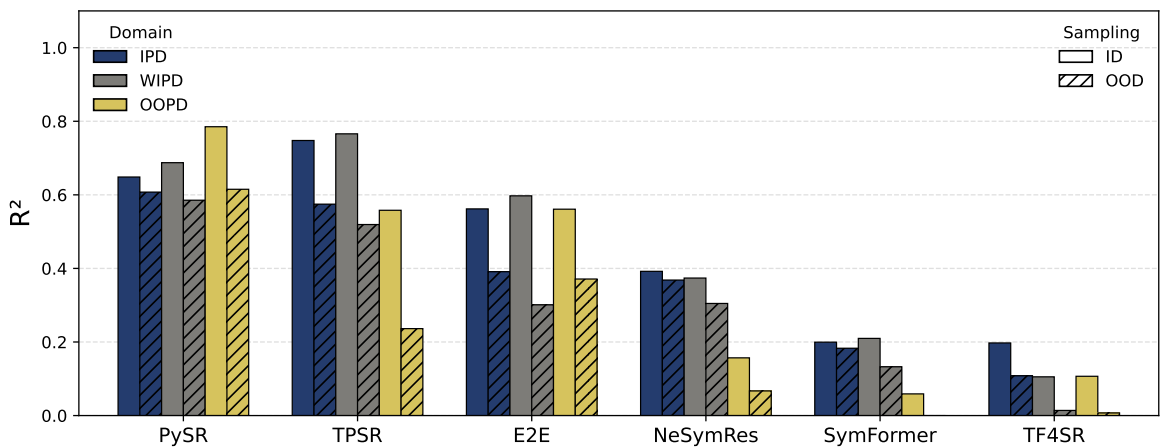

Figure 2: **Predictive accuracy ($R^2$) across domains and sampling regimes.** The figure evaluates the coefficient of determination ($R^2$, higher is better) on the different domain generalization scenarios. Performance is shown for two test set types: in-domain (**ID**, solid bars) and out-of-domain (**OOD**, hatched bars). This evaluation is further broken down by the training data's relation to the model's pre-training domain: in-pre-training (**IPD**), within-pre-training (**WIPD**), and out-of-pre-training (**OOPD**), indicated by the bar colors. The plot reveals that pre-trained methods suffer a significant accuracy drop both when tested on OOD data and when trained on data outside their pre-training domain. In contrast, the search-based PySR demonstrates substantially more robust performance across all conditions.

The hybrid TPSR method further illustrates this dependency. While its ID accuracy is high, it also experiences a significant drop on OOD data (e.g., from 75% to 57% in the IPD case). This gap widens dramatically in the OOPD regime (from 56% to 24%), likely because the E2E prior becomes brittle and provides unreliable guidance to the Monte Carlo Tree Search when operating outside its pre-training domain. Conversely, the search-based methods as a group maintain stable performance with only a slight drop in accuracy, reinforcing the robustness benefits of successful symbolic recovery.

**Complexity results.** The rightmost columns of Table 2 reveal a wide disparity in the complexity of the solutions produced by different methods. We can group the approaches into distinct complexity tiers. At the most concise end, exhaustive search and the NeSymReS transformer consistently produce simple expressions, averaging 4–5 operators. PySR generates moderately more complex solutions, with an average of 7–8 operators. The complexity increases substantially for the hybrid TPSR (13–16 operators) and culminates with E2E, which produces highly complex expressions averaging 20–21 operators. The high complexity of E2E's solutions raises concerns about their interpretability, particularly since other high-performing methods like PySR achieve strong results with vastly simpler expressions. This suggests that E2E's high in-domain accuracy may be an artifact of fitting a large number of free parameters, achieving a numerical fit at the expense of discovering a parsimonious and interpretable symbolic form.

▶ We summarize the outcomes of the generalization experiments as follows:

- **NeSymReS and SymFormer reflect successful pre-training strategies.** The relatively strong

IPD and WIPD recovery of NeSymReS can be attributed to its pre-training on a large set of synthetically generated, concise expressions. Its use of `SymPy` for simplification and canonicalization during data generation likely biases the model toward predicting simple, canonical forms, which aligns well with the benchmark's ground truth expressions. This likely also explains the competitive recovery of SymFormer on the 2-variable subset, given its similar pre-training methodology.

- **TF4SR likely susceptibility to distribution shift.** The poor performance of TF4SR serves as a clear illustration of the brittleness of pre-trained models to distribution shifts. TF4SR was pre-trained on a logarithmic sampling distribution, which prioritizes points near the origin, whereas our evaluation uses uniform sampling. We hypothesize that this mismatch between the training and testing data distributions is the primary cause of its low performance across both recovery and accuracy metrics.

- **E2E acts as a high-fidelity curve fitter.** E2E's combination of high in-domain accuracy, poor OOD generalization, and extremely low recovery can be explained by its core design, which functions more as a flexible curve fitter than a parsimonious symbolic discoverer. Its reliance on data standardization forces the model to operate in a normalized space (z-space), requiring it to predict complex expressions with numerous scaling and shifting constants to map back to the original data domain. This effect is compounded by a subsequent optimization step where these constants are fine-tuned with the BFGS algorithm [Nawi et al., 2006]. While this two-stage process explains its strong ID accuracy, it also accounts for its primary weaknesses: (1) extremely high solution complexity (averaging 20 operators); (2) poor OOD generalization, akin to classical regressors that overfit; and (3) a near-zero recovery rate, as the training objective incentivizes numerical fit over symbolic parsimony.

- **TPSR is bottlenecked by its search prior.** The hybrid TPSR method illustrates both the potential and the critical weakness of combining pre-trained models with search. By guiding its MCTS with a parsimony constraint, TPSR favors shorter expressions, explaining why it achieves a moderate recovery rate while its E2E prior alone recovers none. However, the method's performance is ultimately bottlenecked by the quality of this prior. When the E2E model operates outside its pre-training domain (OOPD), its guidance becomes unreliable. In the vast combinatorial search space of symbolic regression, an unreliable prior renders the search ineffective, explaining TPSR's sharp decline in both recovery and accuracy. This failure is so significant that in OOPD scenarios, its performance can even fall below that of a simple exhaustive search baseline.

## 1.3 Performance Analysis by Problem Complexity

To investigate the impact of problem complexity on model performance, we stratify the SRBench benchmark into three tiers: **easy** ($\leq 4$ operators), **medium** (5–8 operators), and **hard** ($> 8$ operators). The 'easy' tier corresponds to formulas solvable via exhaustive search. To isolate the effect of structural complexity from domain shift, we conduct this analysis exclusively using data from each model's in-pre-training domain (IPD). This allows us to assess the models' capabilities under the most favorable data conditions.

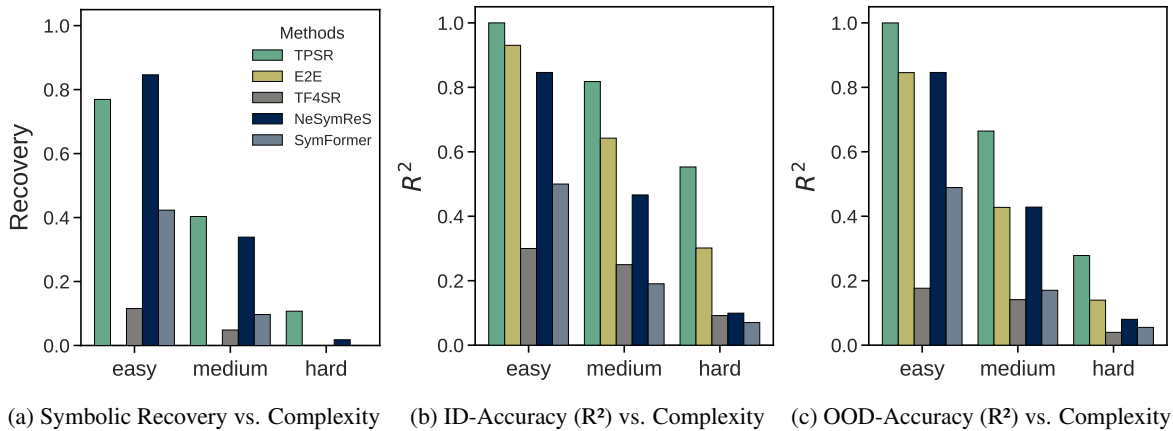

(a) Symbolic Recovery vs. Complexity  (b) ID-Accuracy (R²) vs. Complexity  (c) OOD-Accuracy (R²) vs. Complexity

Figure 3: **Performance stratified by the complexity of the ground-truth symbolic expressions.** The figure presents (a) symbolic recovery rates, (b) in-domain (ID) accuracy ($R^2$), and (c) out-of-domain (OOD) accuracy ($R^2$), stratified into three complexity tiers—easy, medium, and hard—based on the structure of the SRBench formulas. A clear trend is observable across all methods and metrics: performance consistently and significantly degrades as the complexity of the target formula increases. While methods like NeSymReS and TPSR show strong performance on 'easy' problems, their effectiveness diminishes rapidly on 'medium' and 'hard' instances, highlighting that solving structurally complex symbolic regression problems remains a major challenge for current pre-trained and hybrid models.

The results, shown in Figure 3, reveal a consistent and significant performance degradation for all methods as problem complexity increases. This trend is evident in both recovery and accuracy metrics and is expected given the combinatorial explosion of the search space for more complex formulas. In terms of **recovery**, TPSR and NeSymReS emerge as the top performers, solving nearly $80\%$ of 'easy' problems. However, their success rate halves to around $40\%$ on 'medium' problems and drops near zero for 'hard' ones. SymFormer performs moderately on 'easy' problems ( $40\%$) before declining sharply, while E2E and TF4SR exhibit low recovery rates across all complexity levels. Regarding **accuracy** (ID and OOD), a similar pattern holds. TPSR, E2E, and NeSymReS perform strongly on the 'easy' tier, but their accuracy erodes as complexity grows. SymFormer and TF4SR occupy the lower performance tiers. This analysis underscores that even within their ideal pre-training domains, current models struggle to generalize to structurally complex problems.

### 1.4 Analyzing High Accuracy Performance of E2E

To test the hypothesis that E2E's high accuracy stems from its flexibility rather than symbolic correctness, we designed an experiment using the Feynman dataset. We created subsets of problems with 2, 3, and 4 dimensions (N=14, 33, and 27, respectively). For each dimensionality, we first obtained a single symbolic 'skeleton' by running E2E on one randomly chosen problem. We then tested the flexibility of this fixed skeleton by refitting its free constants to every *other* problem of the same dimensionality, using their respective data. This approach isolates the expressive power of the symbolic structure itself.

The results, shown in Table 3, are striking. The E2E skeletons are highly parameterized, containing 12–18 free constants. Remarkably, by only refitting these constants, a single, fixed skeleton can achieve high average R² scores across entirely different physical problems (e.g., 0.989 for 2D problems). This

9

performance is comparable to, and in some cases surpassed by, a simple polynomial of degree 2 or 3, which often uses a similar or even smaller number of parameters. As a further baseline, we report the best-fitting symbolic expression without any free parameters, which was found by an exhaustive search. Table 4 illustrates the behavior for the top-10 best fitted expressions on the 4D subset with the given skeleton.

This experiment provides strong evidence that E2E functions as a high parameterized curve fitter. Its success is not necessarily from discovering the correct underlying structure, but from generating a highly flexible symbolic scaffold that can be effectively fit to the data via its BFGS optimization step. This underscores the critical importance of enforcing parsimony and using metrics that evaluate symbolic equivalency (e.g., recovery rate, TED or JI) rather than relying solely on predictive accuracy.

Table 3: **Flexibility of a single E2E skeleton compared to baselines.** The 'E2E Skeleton' row shows the average R² achieved when a single, fixed symbolic expression (obtained from one problem) is refit to all other problems of the same dimensionality from the Feynman dataset. This tests the expressive power of the skeleton's structure. We report the number of free constants (# consts) and the average R² over all problems in the set (e.g., N=14 for 2D). The results demonstrate that a single, highly parameterized E2E skeleton can act as a universal approximator, achieving high accuracy. However, often even a simpler polynomial model achieves comparable or superior performance with fewer free parameters. This explains why E2E generally achieves high accuracy: it produces highly parameterised solutions that are often overly complex.

| Method | 2D (N=14) | | 3D (N=33) | | 4D (N=27) | |
|---|---|---|---|---|---|---|
| | # consts | Avg $R^2$ | # consts | Avg $R^2$ | # consts | Avg $R^2$ |
| E2E Skeleton | 12 | 0.989 | 18 | 0.736 | 12 | 0.679 |
| Poly (deg=1) | 3 | 0.855 | 4 | 0.679 | 5 | 0.656 |
| Poly (deg=2) | 6 | 0.981 | 10 | 0.882 | 15 | 0.891 |
| Poly (deg=3) | 10 | 0.996 | 20 | 0.929 | 35 | 0.949 |
| Exhaustive Search | 0 | 0.760 | 0 | 0.782 | 0 | 0.747 |

Table 4: **Qualitative examples of a single E2E skeleton fitting diverse 4D physical formulae.** The complex expression shown at the top is the single, fixed symbolic 'skeleton' produced by E2E. Each row lists a different ground-truth formula from the 4D Feynman dataset. The R² score is achieved by optimizing the skeleton's 12 free constants ($c$) to fit the data generated by the corresponding target formula. The consistently high R² values across fundamentally different physical laws demonstrate that the skeleton's complexity allows it to function as a powerful universal approximator, rather than capturing any true underlying symbolic structure. The fitted constant vector ($c$) changes dramatically for each problem, further highlighting this curve-fitting behavior.

$$c_{11}x_3 + c_3 + c_7\left(c_0 + c_8 x_2\right)\left(c_2 + c_4\left(c_1 x_1 + c_9\right)^{c_7}\right)\left(c_{10} x_0 + c_5 x_3 + c_6\right)$$

| Name | Expression | $R^2$ | $c$ |
|---|---|---|---|
| I.15.3t | $\dfrac{-\frac{x_0 x_2}{x_1} + x_3}{\sqrt{1 - \frac{x_2^2}{x_1^2}}}$ | 0.99882 | $[0.0, 0.0, 0.0, -0.1, -0.3, -0.6, -0.9, -1.0, 0.0, 0.0, 0.9, 1.0]$ |
| III.4.33 | $\dfrac{x_0 x_1}{2\pi\left(e^{\frac{x_0 x_1}{2\pi x_2 x_3}} - 1\right)}$ | 0.99944 | $[-0.6, 0.0, -5.1, -0.5, -9.3, 0.4, 0.0, 2.3, 0.0, 1.9, 0.0, -25.4]$ |
| I.15.3x | $\dfrac{x_0 - x_1 x_3}{\sqrt{-\frac{x_1^2}{x_2^2} + 1}}$ | 0.99208 | $[0.1, 0.0, 5.6, -15.9, -0.4, -62.3, -17.4, 1.6, 0.0, 8.0, -1.1, -58.5]$ |
| III.10.19 | $x_0\sqrt{x_1^2 + x_2^2 + x_3^2}$ | 0.98284 | $[0.9, 0.0, 1.0, -13.7, 0.4, -0.2, 1.0, 5.7, 0.1, 0.9, 0.6, 3.2]$ |
| II.11.27 | $\dfrac{x_0 x_1 x_2 x_3}{-\frac{x_0 x_1}{3} + 1}$ | 0.97717 | $[0.0, -0.1, -8.4, 0.0, 0.8, -0.2, 0.4, -7.7, 0.1, 0.7, -0.8, 0.0]$ |
| I.13.4 | $\dfrac{x_0\left(x_1^2 + x_2^2 + x_3^2\right)}{2}$ | 0.96017 | $[0.2, 0.1, -6.2, -34.1, -0.1, 0.0, -0.3, 5.5, 0.1, 1.5, -0.7, 8.8]$ |
| I.18.4 | $\dfrac{x_0 x_2 + x_1 x_3}{x_0 + x_1}$ | 0.95765 | $[-1.5, 0.4, -0.6, 1.5, 1.7, 0.0, -2.3, -0.4, 0.5, 1.4, -0.7, 0.5]$ |
| I.24.6 | $\dfrac{x_0 x_3^2\left(x_1^2 + x_2^2\right)}{4}$ | 0.93664 | $[-1.6, -0.1, 0.0, 6.6, 0.9, 2.7, -4.3, -3.2, -1.1, 1.3, 1.0, -8.3]$ |
| II.38.3 | $\dfrac{x_0 x_1 x_3}{x_2}$ | 0.91797 | $[1.7, 1.3, 0.7, 0.1, 1.1, 0.7, -2.1, 1.1, -0.3, -0.1, 0.7, 0.1]$ |
| I.38.12 | $\dfrac{x_2^2 x_3}{\pi x_0 x_1^2}$ | 0.91747 | $[5.2, 2.0, 0.0, 0.1, -2.0, -1.3, -3.6, -3.1, -4.0, 1.3, 1.3, 0.0]$ |

# 2 Ablation Study

This section elaborates on the experimental results of an ablation study. First, in section 2.1 we provide details on the experimental setup. Section 2.2 provides a comprehensive breakdown of ablation study findings, including a detailed table of all pre-training configurations, analyses of recovery performance across varying expression lengths, dimensions, and dimension-length $(D, L)$ buckets, and additional metrics on the Feynman dataset. Section 2.3 discusses how the model performance improves when scaling up model parameters. Section 2.4 then contextualizes the best-performing transformer model configuration of our ablation studies within the state-of-the-art on the SRBench benchmark.

## 2.1 Experimental Setup

We have seen that both the recovery and the accuracy performance of the four evaluated transformer-based approaches can differ significantly. To better understand these differences, we conducted an ablation study on how different pre-training parameters, namely, the selection of pre-training expressions and the sampling strategy, impact the performance of the transformer approach in different training domains. To this end, we have implemented our own transformer-based symbolic regression model that follows the architecture of the E2E approach by Kamienny et al. [2022]: Our model has 86 million parameters and uses an embedding dimension of $512$. The point-encoder is a standard 4-layer transformer-encoder, and the decoder has 16 layers. However, in contrast to Kamienny et al. [2023], our approach is not end-to-end, because in the training phase expressions are generated that can contain placeholders for constants. These constants are fitted using the Broyden–Fletcher–Goldfarb–Shanno (BFGS) algorithm [Nawi et al., 2006]. We used this particular implementation to explore different pre-training strategies for transformer-based symbolic regression algorithms. The pre-training and training for all experiments were conducted on a single NVIDIA RTX A6000 GPU.

### 2.1.1 Pre-Training Configurations

We have varied the three pre-training parameters, namely, the selection of symbolic formulas, standardization, and the sampling strategy for these formulas.

- **Selection of symbolic formulas.** We have compared the random sampling of expression trees that is based on the framework by Lample and Charton [2019] to the approach of Kahlmeyer et al. [2024] that systematically enumerates small, semantically different expression DAGs. That is, we have two options for the selection of symbolic formulas, namely, *random expression trees* and *small expression DAGs*.

- **Standardization of inputs.** We compare two pre-training setups: one which first standardizes all inputs before referring them to the model and another one which uses the raw inputs instead. That is, we have two options for standardization: *standardized* and *non-standardized*.

- **Pre-training sampling distribution.** We consider three sampling distributions, namely, uniform, Gaussian and diverse [Kamienny et al., 2022] on the pre-training sampling domains. That is, we have two options for the sampling distribution, namely, *uniform* and *diverse*.

During all pre-training configurations we add Gaussian noise to the output data points as described in [Kamienny et al., 2022]. For the experiments we define the following base configuration: Symbolic

formulas (*expression DAGs*), *non-standardization*, and *diverse* sampling distribution. For each ablation, we retrain a new model from scratch, always changing only one parameter of the base configuration. We report results for all configurations on all three evaluation regimes IPD, WIPD, and OOPD.

### 2.1.2 Training Hyperparameters

Table 5 contains the hyperparameters used to train the transformer model in the ablation studies. The model is trained by minimizing a cross-entropy loss using the Adam optimizer. The learning rate is initially set to $10^{-7}$ and then, over the first 2000 steps, warmed up to $10^{-4}$. Subsequently, a cosine learning rate decay to $10^{-5}$ is employed, as in [Lewkowycz, 2021].

| Parameter | Value |
| --- | --- |
| Embedding Dimension | 512 |
| Num Encoder Layers | 4 |
| Num Decoder Layers | 1, 2, 4, 8, 16 |
| Num Attention Heads | 16 |
| Batch Size | 32 |
| Learning Rate | $10^{-4}$ |
| Warmup Iterations | 2000 |
| Weight Decay | $10^{-1}$ |
| Adam Betas | (0.9, 0.95) |
| Dropout | 0.0 |
| Grad Clip | 0.5 |

Table 5: **Transformer model and training hyperparameters.** This table lists the specific values for the architectural and training hyperparameters of the transformer model used in this study.

## 2.2 Results

Table 6 and Table 7 presents the complete results of all experiments conducted in the ablation study, evaluating the impact of various pre-training configurations on transformer-based symbolic regression performance. The configurations tested include: (I) expression selection (small expression DAGs, random expression trees), (II) point sampling strategy during pre-training (uniform, diverse), and (III) input standardization (with and without standardization). All configurations were evaluated on the SRBench dataset [La Cava et al., 2021], using test data points generated with three distinct sampling distributions: uniform, Gaussian, and diverse.

Examination of the results reveals that models achieve strong recovery performance when the test distribution resembles the point sampling distribution used during pre-training. However, recovery performance degrades significantly when the pre-training distribution differs from the test distribution. Notably, Gaussian sampling during testing consistently yields the highest recovery rates across all ablation configurations. This suggests that sampling points close to the origin provides valuable information for formula recovery. Furthermore, the absence of succinct expressions during pre-training

Table 6: **Ablation study on pre-training data generation choices and their impact on symbolic recovery.** We compare a **Base** model (pre-trained on enumerated expression DAGs with diverse point sampling) against three single-variation ablations: **+ Tree** (using randomly sampled expression trees instead of DAGs), **+ Uniform** (using uniform point sampling instead of diverse), and **+ Std** (adding input data standardization). The entire experiment is repeated for three distinct test distributions: Uniform, Gaussian, and Diverse. Performance is evaluated across in-pre-training (IPD), within-pre-training (WIPD), and out-of-pre-training (OOPD) domains using Recovery Rate (higher is better), Tree Edit Distance (lower is better), and Jaccard Index (higher is better). The results show that the **Base** model's configuration is generally superior, especially in-domain. While standardization (**+ Std**) offers some robustness in OOPD scenarios, altering the expression generation (**+ Tree**) or simplifying the sampling distribution (**+ Uniform**) significantly degrades performance.

| Method | Recovery | | | Tree Edit Distance | | | Jaccard Index | | |
| --- | --- | --- | --- | --- | --- | --- | --- | --- | --- |
| | IPD | WIPD | OOPD | IPD | WIPD | OOPD | IPD | WIPD | OOPD |
| **Test Distribution: Uniform** | | | | | | | | | |
| Base | 0.35 | 0.37 | 0.09 | 0.67 | 0.60 | 0.84 | 0.35 | 0.44 | 0.22 |
| + Tree | 0.03 | 0.03 | 0.03 | 0.94 | 0.94 | 0.95 | 0.21 | 0.21 | 0.17 |
| + Uniform | 0.35 | 0.00 | 0.01 | 0.61 | 0.95 | 0.95 | 0.44 | 0.15 | 0.17 |
| + Std | 0.23 | 0.26 | 0.23 | 0.73 | 0.68 | 0.72 | 0.36 | 0.38 | 0.35 |
| **Test Distribution: Gaussian** | | | | | | | | | |
| Base | 0.40 | 0.35 | 0.09 | 0.59 | 0.63 | 0.89 | 0.45 | 0.42 | 0.20 |
| + Tree | 0.03 | 0.03 | 0.00 | 0.94 | 0.94 | 0.97 | 0.22 | 0.21 | 0.17 |
| + Uniform | 0.19 | 0.00 | 0.00 | 0.80 | 0.98 | 0.98 | 0.26 | 0.05 | 0.06 |
| + Std | 0.25 | 0.26 | 0.17 | 0.70 | 0.66 | 0.76 | 0.38 | 0.41 | 0.32 |
| **Test Distribution: Diverse** | | | | | | | | | |
| Base | 0.24 | 0.16 | 0.03 | 0.71 | 0.73 | 0.89 | 0.35 | 0.33 | 0.20 |
| + Tree | 0.03 | 0.03 | 0.00 | 0.94 | 0.94 | 0.98 | 0.21 | 0.19 | 0.14 |
| + Uniform | 0.00 | 0.01 | 0.00 | 0.97 | 0.98 | 0.97 | 0.06 | 0.07 | 0.08 |
| + Std | 0.15 | 0.17 | 0.16 | 0.74 | 0.75 | 0.79 | 0.34 | 0.33 | 0.25 |

consistently results in low recovery rates. Standardization of inputs leads to a noticeable improvement in OOPD performance compared to non-standardized configurations, but at the cost of reduced IPD and WIPD recovery. Tree Edit Distance (TED), a more nuanced metric that quantifies the distance to the original formula in terms of tree edits, supports these findings. Models trained using small expression DAGs and diverse sampling show superior performance on the TED metric, indicating consistency with the recovery rate results. In terms of model fit, models generally achieve good IPD and WIPD fit, but OOPD fit is significantly worse. Finally, models pre-trained on succinct expressions generate significantly less complex formulae while maintaining or even improving accuracy, highlighting a potential advantage in terms of interpretability.

▶, We summarize our findings as follows:

- **Impact of the selection of the symbolic expressions in pre-training.** Our base model, pre-trained on systematically enumerated expression DAGs and diverse samplings, achieves a recovery rate of $0.35$ on uniformly sampled IPD test data. Switching from DAGs to randomly

Table 7: **Ablation study on the impact of pre-training choices on predictive accuracy (R²) and complexity.**
We compare a **Base** model (pre-trained on enumerated DAGs with diverse point sampling) against three variations:
expression source (**+ Tree**), point sampling (**+ Uniform**), and data preprocessing (**+ Std**). The entire experiment
is repeated for three distinct **Test Distributions** (Uniform, Gaussian, Diverse). Within each section, we evaluate
across three training domains (IPD, WIPD, OOPD) and report R² (higher is better) for both **in-domain (ID)** and
**out-of-domain (OOD)** test sets, as well as solution **complexity**. The results highlight several key trade-offs:
the **Base** model excels in-domain, while standardization (**+ Std**) improves OOD robustness at the cost of peak
performance. Notably, pre-training on random trees (**+ Tree**) results in significantly more complex and less
accurate solutions, underscoring the importance of a curated, concise pre-training set.

|  | Accuracy-ID (R²) | | | Accuracy-OOD (R²) | | | Complexity | | |
|---|---|---|---|---|---|---|---|---|---|
| **Method** | **IPD** | **WIPD** | **OOPD** | **IPD** | **WIPD** | **OOPD** | **IPD** | **WIPD** | **OOPD** |
| **Test Distribution: Uniform** | | | | | | | | | |
| Base | 0.53 | 0.59 | 0.27 | 0.48 | 0.48 | 0.17 | 6.21 | 6.99 | 6.74 |
| + Tree | 0.13 | 0.18 | 0.19 | 0.10 | 0.08 | 0.07 | 11.33 | 11.28 | 11.40 |
| + Uniform | 0.56 | 0.15 | 0.24 | 0.52 | 0.08 | 0.03 | 6.40 | 9.10 | 9.18 |
| + Std | 0.42 | 0.47 | 0.48 | 0.38 | 0.39 | 0.35 | 6.89 | 6.64 | 6.78 |
| **Test Distribution: Gaussian** | | | | | | | | | |
| Base | 0.60 | 0.62 | 0.32 | 0.57 | 0.45 | 0.19 | 7.23 | 7.20 | 7.17 |
| + Tree | 0.10 | 0.22 | 0.23 | 0.09 | 0.04 | 0.06 | 11.16 | 10.99 | 11.03 |
| + Uniform | 0.32 | 0.05 | 0.15 | 0.27 | 0.01 | 0.01 | 7.76 | 10.71 | 9.23 |
| + Std | 0.46 | 0.54 | 0.47 | 0.40 | 0.39 | 0.29 | 6.95 | 6.65 | 7.10 |
| **Test Distribution: Diverse** | | | | | | | | | |
| Base | 0.50 | 0.45 | 0.17 | 0.34 | 0.23 | 0.08 | 7.22 | 7.58 | 7.07 |
| + Tree | 0.19 | 0.18 | 0.12 | 0.05 | 0.04 | 0.02 | 10.86 | 11.85 | 11.42 |
| + Uniform | 0.09 | 0.13 | 0.10 | 0.01 | 0.01 | 0.00 | 10.11 | 9.63 | 10.17 |
| + Std | 0.49 | 0.44 | 0.35 | 0.27 | 0.26 | 0.26 | 6.81 | 6.84 | 7.02 |

sampled expression trees results in a significant drop in recovery: $0.03$ under a diverse test
distribution and uniform test distribution, respectively. On WIPD and OOPD data, models also
exhibit similar drops in recovery, albeit from a lower initial baseline compared to IPD. We also
observe performance drops in terms of accuracy, $R^2$ score, on model configurations that have
been evaluated on IPD and WIPD data.

- **Impact of the sampling distribution.** Across all three domains (IPD, WIPD, and OOPD),
  models pre-trained with a diverse sampling distribution exhibit only a slight decrease in accuracy
  when the test distribution shifts from diverse to uniform. In contrast, models pre-trained with
  a uniform sampling distribution experience a significant drop in accuracy when evaluated on
  a diverse test distribution. This pattern is mirrored in the recovery metric: models pre-trained
  on diverse distributions maintain stable recovery under uniform test conditions, whereas the
  recovery of models pre-trained on uniform distributions declines substantially when tested with a
  diverse distribution.

- **Impact of the sampling domain.** As expected the base model's recovery performance declines

when moving the training data from IPD to WIPD, and breaks down when moving to OOPD training data. This is similar to the accuracy performance, which also degrades when moving from IPD to WIPD to OOPD. It is worth noting that models pre-trained on standardized input data show lower overall performance, but are more robust when transitioning from IPD to WIPD to OOPD.

## 2.3 Scaling Ablations

As the ablations in Section 2.2 have shown, the performance of the best performing Base model decreases when moving out of the pre-training domain. Nevertheless, a pre-trained model can be fairly general, if the pre-training domain is large enough. This section explores how scaling key parameters influences IPD recovery performance. We systematically ablate model size (number of parameters), pre-training dataset size, and beam size during inference to assess their impact on IPD recovery. Results, evaluated on 119 formulas from the Feynman dataset [Feynman et al., 2011], are presented in Figure 4.

**Impact of model size.** We observe a linear improvement in IPD recovery performance, from 0.4 to 0.5, as the number of decoder layers in the transformer model - and consequently, the total number of parameters - increases (Figure 4). This suggests that IPD recovery can be enhanced by scaling up model capacity. However, the linear nature of this improvement implies that achieving performance comparable to search-based symbolic regression methods would likely incur substantial computational costs due to the required model scale. In Table 8, we present additional results for the scaling experiments over the parameter size of the model. The parameter size is increased by scaling the number of decoder layers from one to 16. We report recovery, $R^2$-score, Root Mean Squared Error (RMSE), and complexity on the Feynman data set. Complexity is measured as the number of nodes in the expression tree of the generated symbolic regressor, following [La Cava et al., 2021]. In addition, we report recovery performance on training and test expressions. Larger parameter sizes consistently show higher recovery rates.

**Impact of dataset size and beam size.** For the largest model variant with 16 decoder layers, we perform ablations over dataset size and beam size. In the dataset size experiments, we scale the number of distinct expressions a model sees during training from one to 16 million. We train each model until the validation loss is saturated and then measure the recovery performance. As shown in Figure 4, regardless of the number of model parameters, we observe a stable increase in recovery performance as more distinct expressions are seen during training. This strong correlation underscores the importance of diverse and extensive pre-training data; more unique, succinct representations seen during training directly translate to better IPD recovery. Similar to model size, this indicates that IPD performance gains can be achieved through increased computational investment in data generation. Given the synthetic nature of the data generation method, we expect the in-pre-training-domain recovery performance to increase further as more synthetic data is generated. In the beam size experiments, we infer the model with varying beam sizes from 1 to 16. We observe that employing a beam search significantly boosts recovery. Performance improvements saturate around a beam size of 16, suggesting diminishing returns and limited benefit from investing further computational resources into wider beam searches at test time.

▶ In summary, our scaling ablations demonstrate that while improvements in IPD recovery are achievable, they necessitate significant computational investment. However, the performance of these

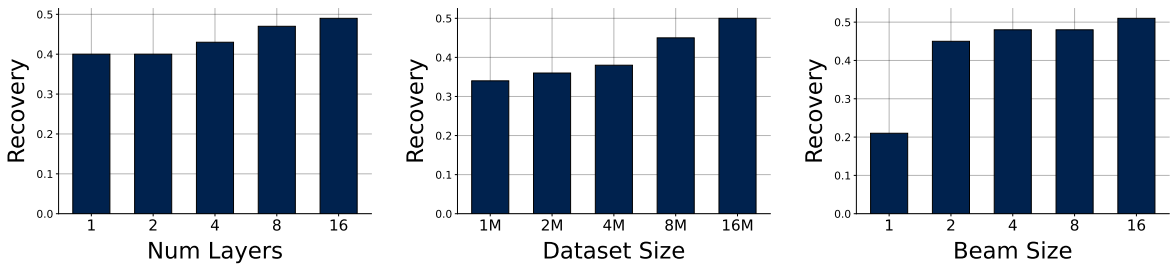

Figure 4: **Scaling laws for IPD symbolic recovery performance.** The figure investigates the impact of three key scaling factors on the in-pre-training-domain (IPD) recovery rate. We vary **(a)** the number of transformer layers (model depth), **(b)** the size of the pre-training dataset (in millions of distinct expressions), and **(c)** the beam size used during inference. The results show a strong, positive correlation between dataset size and performance and highlight the critical importance of beam search, with performance jumping significantly from a beam size of 1 to 2. Increasing model depth also improves recovery, though the gains are less pronounced. This analysis suggests that scaling the pre-training dataset is the most effective strategy for improving in-domain performance, while even minimal search at inference is essential.

| Model | Recov Feynman | R² | RMSE | Complexity | Recov Test | Recov Train |
|---|---|---|---|---|---|---|
| TF-1  | 0.40 | 0.83 | 0.09 | 10.7 | 0.09 | 0.11 |
| TF-2  | 0.40 | 0.86 | 0.09 | 11.0 | 0.10 | 0.11 |
| TF-4  | 0.43 | 0.85 | 0.07 | 11.4 | 0.10 | 0.13 |
| TF-8  | 0.47 | 0.85 | 0.06 | 11.4 | 0.12 | 0.14 |
| TF-16 | 0.49 | 0.87 | 0.07 | 11.2 | 0.13 | 0.15 |

Table 8: **Impact of model depth on performance.** We evaluate transformer models (**TF-L**) with a varying number of decoder layers ($L \in \{1, 2, 4, 8, 16\}$). Performance is assessed on four distinct metrics: recovery rate, R², RMSE, complexity). We report recovery on the Feynman dataset, a **held-out test set** of synthetic expressions from the pre-training distribution ('Recov Test'), and the **synthetic training set** itself ('Recov Train'). The results demonstrate a clear and positive scaling law: increasing model depth consistently improves recovery rates on all three datasets without a significant increase in solution complexity.

ablations remains below that of state-of-the-art search-based methods.

**Roles of memorization and generalization in expression recovery.** To estimate the ability of the model to generalize to unseen expressions, we need to consider the overlap between the Feynman dataset and the pre-training data. Approximately $51\%$ of the Feynman expressions are present in the pre-training dataset. Consequently, successful recovery on this dataset likely reflects a blend of memorization and generalization. Further analysis reveals that $82.6\%$ of the formulas recovered from the Feynman dataset were present in the pre-training data, suggesting they were successfully memorized. Conversely, the remaining $17.4\%$ were not in the pre-training data, indicating that their recovery is attributable to the model's generalization ability. It is important to note that memorization in this context is not simply rote learning of input-output pairs. Due to the highly diverse sampling strategy employed during pre-training, it is unlikely that the model has encountered the exact test instances

17

during pre-training. Our findings indicate that while a substantial portion of successfully recovered expressions (82.6%) can be attributed to memorization, a non-negligible fraction (17.4%) demonstrates generalization to previously unseen expressions. This observation is supported by the performance on the training and test expressions. Specifically, our ablation studies varying the number of decoder layers demonstrate that the capacity for IPD generalization, though initially limited, improves with increasing model parameter count. This trend aligns with the findings of Kaplan et al. [2020], who reported a log-linear relationship between model size and generalization performance on language modeling tasks. A similar relationship can be observed for our task.

| Ablation | Recov Feynman | $R^2$ | RMSE | Complexity |
|---|---|---|---|---|
| **Dataset Size** | | | | |
| TF-16 1M | 0.34 | 0.79 | 0.07 | 11.72 |
| TF-16 2M | 0.36 | 0.76 | 0.07 | 12.86 |
| TF-16 4M | 0.38 | 0.79 | 0.07 | 12.06 |
| TF-16 8M | 0.45 | 0.79 | 0.06 | 11.51 |
| TF-16 16M | 0.50 | 0.88 | 0.07 | 11.30 |
| **Beam Size** | | | | |
| TF-16 1BS | 0.21 | 0.56 | 0.05 | 17.68 |
| TF-16 2BS | 0.45 | 0.76 | 0.05 | 12.28 |
| TF-16 4BS | 0.48 | 0.80 | 0.04 | 11.52 |
| TF-16 8BS | 0.48 | 0.84 | 0.07 | 11.16 |
| TF-16 16BS | 0.51 | 0.91 | 0.06 | 10.97 |

Table 9: **Impact of dataset size and beam size on performance for the TF-16 model.** We ablate two key parameters for the 16-layer transformer model (TF-16) on the Feynman benchmark: the pre-training dataset size (1M to 16M expressions) and the beam size used during inference (1 to 16). The results show a strong positive correlation between **dataset size** and performance, with recovery increasing from 0.34 to 0.50. The impact of **beam size** is even more pronounced: moving from greedy decoding (beam size 1) to a minimal search (beam size 2) more than doubles the recovery rate (0.21 to 0.45) and substantially reduces the complexity of the generated expressions (17.68 to 12.28). This analysis highlights that while larger datasets are crucial, even a small amount of search during inference is essential for both accuracy and parsimony.

**Recovery per length and dimension.** Henceforth, we will use *length* of an expression to denote the number of operators in its string, as counted in SRBench [La Cava et al., 2021]. Figure 5 illustrates the recovery performance of the model on the Feynman dataset over different expression lengths and dimensions. It can be observed that expressions with shorter lengths and expressions with fewer input dimensions are generally recovered better than longer expressions or expressions with a larger number of inputs. This is not a phenomenon specific to the transformer architecture. Furthermore, the steady decline in recovery performance can be observed for various state-of-the-art symbolic regression methods such as UDFS [Kahlmeyer et al., 2024], DSR [Landajuela et al., 2021] or PySR [Cranmer, 2023], as shown in Figure 6. Note that complexity here is usually measured as the size of the minimal expression tree of the target expression to be recovered.

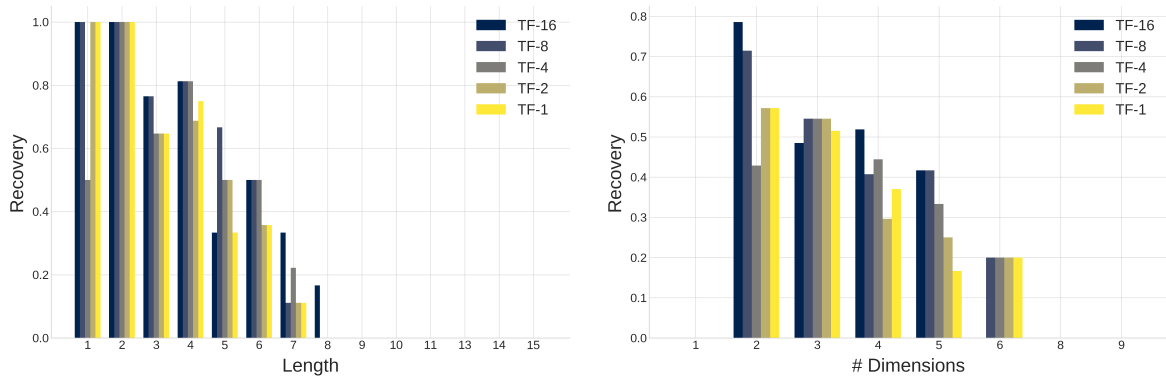

Figure 5: **Detailed analysis of recovery performance on the Feynman dataset.** The figure illustrates recovery performance metrics: (Left) stratified by expression length in terms of number of operators, and (Right) stratified by the number of input dimensions (variables) of the target expressions.

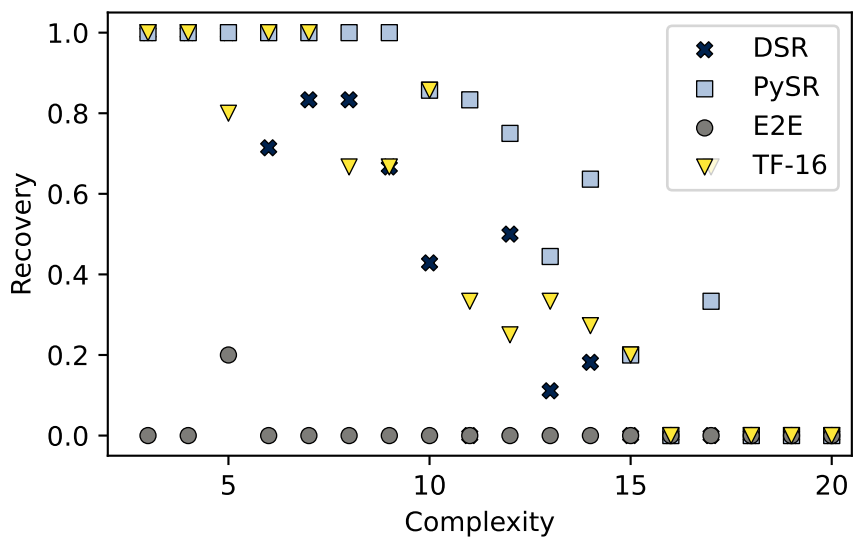

Figure 6: **Comparison of recovery over complexity for state-of-the-art symbolic regression methods.** The recovery performance steadily decreases as the formulas to be recovered become more complex. This is a phenomenon observed for several state-of-the-art symbolic regression methods, such as PySR [Cranmer, 2023] or DSR [Petersen et al., 2021].

**Recovery over dimension-length buckets.** Figure 7 illustrates the recovery performance of model ablations with different parameter sizes over all $(D, L)$ buckets. Models with a larger number of parameters consistently achieve higher recovery across all buckets.

**Inference speed comparison.** To put the inference speeds of the various state-of-the-art symbolic regression methods in the context of the pre-trained, transformer-based approach, we measure the performance of the TF-16 transformer model configuration on our setup with a single NVIDIA RTX A6000 GPU. Table 10 shows the results of the comparison. Compared to other methods, speedups of

19

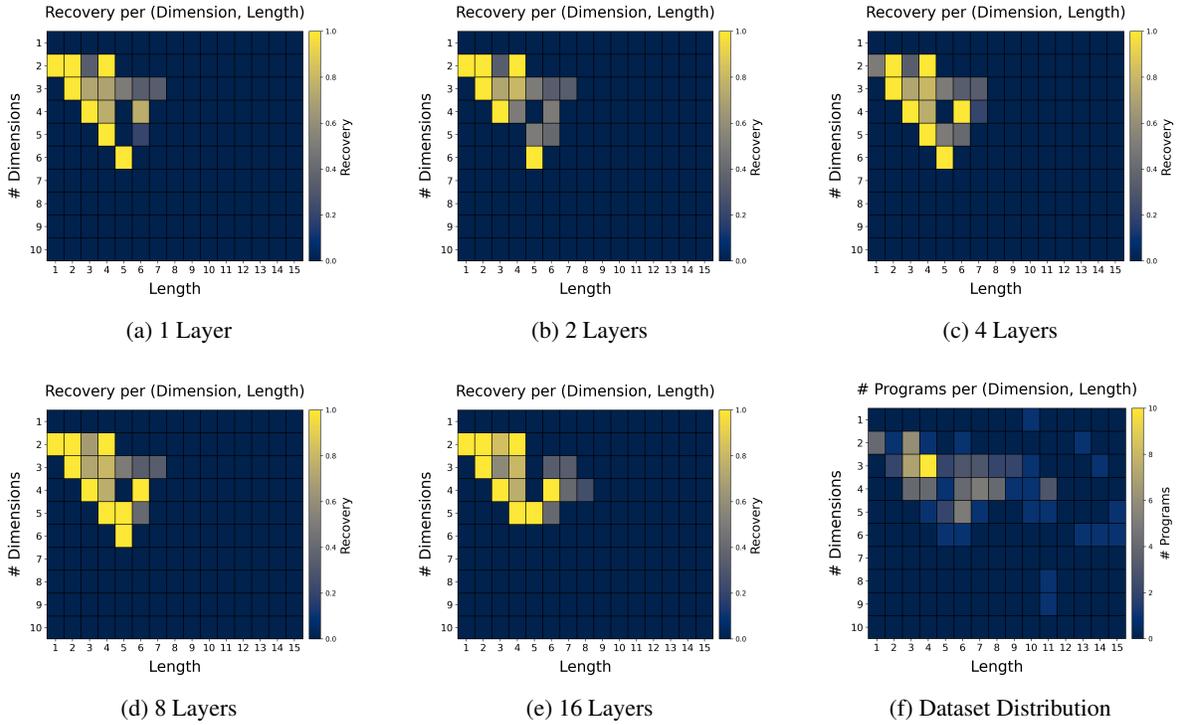

Figure 7: **Distribution of recovery results over $(D, L)$ buckets.** Distribution of recovery results for different model sizes over the different combinations of length and input dimension of the Feynman data set. The plot in the lower right-hand corner shows the overall distribution of the test expressions in the data set.

2.3 to 3 orders of magnitude (depending on the inference platform) are achieved.

**Detailed analysis of the formulas recovered from the Feynman data set.** The Feynman problems comprise formulas from the renowned Feynman Lectures Feynman et al. [2011] and were initially introduced by Udrescu and Tegmark [2020] to assess the efficacy of their symbolic regression method. Subsequently, it has been employed and validated by La Cava et al. [2021] in their comprehensive benchmark suite. The database can be accessed via the following link: here.

| Name | Expression | Name | Expression |
| --- | --- | --- | --- |
| I.10.7 | $\frac{x_0}{\sqrt{-\frac{x_1^2}{x_2^2}+1}}$ | I.11.19 | $x_0 x_3 + x_1 x_4 + x_2 x_5$ |
| **I.12.1** | $x_0 x_1$ | **I.12.11** | $x_0 \left( x_1 + x_2 x_3 \sin(x_4) \right)$ |
| **I.12.2** | $\frac{x_0 x_1}{4\pi x_2 x_3^2}$ | I.12.4 | $\frac{x_0}{4\pi x_1 x_2^2}$ |
| **I.12.5** | $x_0 x_1$ | I.13.12 | $x_0 x_1 x_4 \cdot \left( \frac{1}{x_3} - \frac{1}{x_2} \right)$ |
| **I.13.4**[*] | $\frac{x_0 \left( x_1^2 + x_2^2 + x_3^2 \right)}{2}$ | **I.14.3** | $x_0 x_1 x_2$ |
| **I.14.4** | $\frac{x_0 x_1^2}{2}$ | I.15.3t | $\frac{-\frac{x_0 x_2}{x_1^2} + x_3}{\sqrt{1 - \frac{x_2^2}{x_1^2}}}$ |

Continues on next page

20

| Name | Expression | Name | Expression |
| --- | --- | --- | --- |
| I.15.3x | $\frac{x_0 - x_1 x_3}{\sqrt{-\frac{x_1^2}{x_2^2} + 1}}$ | I.16.6 | $\frac{x_1 + x_2}{1 + \frac{x_1 x_2}{x_0^2}}$ |
| I.18.12 | $x_0 x_1 \sin(x_2)$ | I.18.14 | $x_0 x_1 x_2 \sin(x_3)$ |
| **I.18.4** | $\frac{x_0 x_2 + x_1 x_3}{x_0 + x_1}$ | I.24.6 | $\frac{x_0 x_3^2 (x_1^2 + x_2^2)}{4}$ |
| **I.25.13** | $\frac{x_0}{x_1}$ | I.27.6 | $\frac{1}{\frac{x_2}{x_1} + \frac{1}{x_0}}$ |
| I.29.16 | $\sqrt{x_0^2 - 2x_0 x_1 \cos(x_2 - x_3) + x_1^2}$ | **I.29.4** | $\frac{x_0}{x_1}$ |
| I.30.3 | $\frac{x_0 \sin^2\left(\frac{x_1 x_2}{2}\right)}{\sin^2\left(\frac{x_1}{2}\right)}$ | I.32.17 | $\frac{4\pi x_0 x_1 x_2^2 x_3^2 x_4^4}{3(x_4^2 - x_5^2)^2}$ |
| **I.32.5**[*] | $\frac{x_0^2 x_1^2}{6\pi x_2 x_3^3}$ | I.34.1 | $\frac{x_2}{1 - \frac{x_1}{x_0}}$ |
| I.34.14 | $\frac{x_2 \cdot \left(1 + \frac{x_1}{x_0}\right)}{\sqrt{1 - \frac{x_1^2}{x_0^2}}}$ | **I.34.27** | $\frac{x_0 x_1}{2\pi}$ |
| **I.34.8** | $\frac{x_0 x_1 x_2}{x_3}$ | I.37.4 | $x_0 + x_1 + 2\sqrt{x_0 x_1} \cos(x_2)$ |
| **I.38.12**[*] | $\frac{x_2^2 x_3}{\pi x_0 x_1^2}$ | **I.39.1** | $\frac{3 x_0 x_1}{2}$ |
| I.39.11 | $\frac{x_1 x_2}{x_0 - 1}$ | **I.39.22** | $\frac{x_0 x_1 x_3}{x_2}$ |
| I.40.1 | $x_0 e^{-\frac{x_1 x_2 x_4}{x_3 x_5}}$ | I.41.16 | $\frac{x_0^3 x_2}{2\pi^3 x_4^2 \left(e^{\frac{x_0 x_2}{2\pi x_1 x_3}} - 1\right)}$ |
| **I.43.16** | $\frac{x_0 x_1 x_2}{x_3}$ | **I.43.31** | $x_0 x_1 x_2$ |
| I.43.43 | $\frac{x_1 x_3}{x_2 (x_0 - 1)}$ | **I.44.4**[*] | $x_0 x_1 x_2 \log\left(\frac{x_4}{x_3}\right)$ |
| **I.47.23** | $\sqrt{\frac{x_0 x_1}{x_2}}$ | I.50.26 | $x_0 \left(x_3 \cos^2(x_1 x_2) + \cos(x_1 x_2)\right)$ |
| I.6.2 | $\frac{\sqrt{2} e^{-\frac{x_1^2}{2x_0^2}}}{2\sqrt{\pi} x_0}$ | I.6.2a | $\frac{\sqrt{2} e^{-\frac{x_0^2}{2}}}{2\sqrt{\pi}}$ |
| I.6.2b | $\frac{\sqrt{2} e^{-\frac{(x_1 - x_2)^2}{2x_0^2}}}{2\sqrt{\pi} x_0}$ | **I.8.14**[*] | $\sqrt{(-x_0 + x_1)^2 + (-x_2 + x_3)^2}$ |
| I.9.18 | $\frac{x_0 x_1 x_2}{(-x_3 + x_4)^2 + (-x_5 + x_6)^2 + (-x_7 + x_8)^2}$ | **II.10.9** | $\frac{x_0}{x_1 (x_2 + 1)}$ |
| **II.11.20**[*] | $\frac{x_0 x_1^2 x_2}{3 x_3 x_4}$ | II.11.27 | $\frac{x_0 x_1 x_2 x_3}{-\frac{x_0 x_1}{3} + 1}$ |
| II.11.28 | $\frac{x_0 x_1}{-\frac{x_0 x_1}{3} + 1} + 1$ | II.11.3 | $\frac{x_0 x_1}{x_2 (x_3^2 - x_4^2)}$ |
| **II.13.17** | $\frac{x_2}{2\pi x_0 x_1^2 x_3}$ | II.13.23 | $\frac{x_0}{\sqrt{-\frac{x_1^2}{x_2^2} + 1}}$ |
| II.13.34 | $\frac{x_0 x_1}{\sqrt{-\frac{x_1^2}{x_2^2} + 1}}$ | **II.15.4** | $-x_0 x_1 \cos(x_2)$ |
| **II.15.5** | $-x_0 x_1 \cos(x_2)$ | **II.2.42**[*] | $\frac{x_0 x_3 (-x_1 + x_2)}{x_4}$ |
| II.21.32 | $\frac{x_0}{4\pi x_1 x_2 \left(-\frac{x_3}{x_4} + 1\right)}$ | II.24.17 | $\sqrt{\frac{x_0^2}{x_1^2} - \frac{\pi^2}{x_2^2}}$ |
| **II.27.16** | $x_0 x_1 x_2^2$ | **II.27.18** | $x_0 x_1^2$ |
| II.3.24 | $\frac{x_0}{4\pi x_1^2}$ | **II.34.11** | $\frac{x_0 x_1 x_2}{2 x_3}$ |
| II.34.2 | $\frac{x_0 x_1 x_2}{2}$ | **II.34.29a** | $\frac{x_0 x_1}{4\pi x_2}$ |

| Name | Expression | Name | Expression |
|---|---|---|---|
| **II.34.29b**[*] | $\frac{2\pi x_0 x_2 x_3 x_4}{x_1}$ | **II.34.2a** | $\frac{x_0 x_1}{2\pi x_2}$ |
| II.35.18 | $\frac{x_0}{e^{\frac{x_3 x_4}{x_1 x_2}} + e^{-\frac{x_3 x_4}{x_1 x_2}}}$ | II.35.21 | $x_0 x_1 \tanh\left(\frac{x_1 x_2}{x_3 x_4}\right)$ |
| II.36.38 | $\frac{x_0 x_1}{x_2 x_3} + \frac{x_0 x_4 x_7}{x_2 x_3 x_5 x_6^2}$ | **II.37.1** | $x_0 x_1 (x_2 + 1)$ |
| II.38.14 | $\frac{x_0}{2x_1 + 2}$ | **II.38.3** | $\frac{x_0 x_1 x_3}{x_2}$ |
| **II.4.23** | $\frac{x_0}{4\pi x_1 x_2}$ | II.6.11 | $\frac{x_1 \cos(x_2)}{4\pi x_0 x_3^2}$ |
| II.6.15a | $\frac{3 x_1 x_5 \sqrt{x_3^2 + x_4^2}}{4\pi x_0 x_2^5}$ | II.6.15b | $\frac{3 x_1 \sin(x_2) \cos(x_2)}{4\pi x_0 x_3^3}$ |
| **II.8.31** | $\frac{x_0 x_1^2}{2}$ | **II.8.7** | $\frac{3 x_0^2}{20\pi x_1 x_2}$ |
| III.10.19 | $x_0 \sqrt{x_1^2 + x_2^2 + x_3^2}$ | **III.12.43** | $\frac{x_0 x_1}{2\pi}$ |
| **III.13.18** | $\frac{4 x_0 x_1^2 x_2}{x_3}$ | III.14.14 | $x_0 \left(e^{\frac{x_1 x_2}{x_3 x_4}} - 1\right)$ |
| III.15.12 | $2x_0 \cdot (1 - \cos(x_1 x_2))$ | **III.15.14** | $\frac{x_0^2}{8\pi^2 x_1 x_2^2}$ |
| **III.15.27** | $\frac{2\pi x_0}{x_1 x_2}$ | III.17.37 | $x_0 (x_1 \cos(x_2) + 1)$ |
| III.19.51 | $-\frac{x_0 x_1^4}{8 x_2^2 x_3^2 x_4^2}$ | **III.21.20** | $-\frac{x_0 x_1 x_2}{x_3}$ |
| III.4.32 | $\frac{1}{e^{\frac{x_0 x_1}{2\pi x_2 x_3}} - 1}$ | III.4.33 | $\frac{x_0 x_1}{2\pi\left(e^{\frac{x_0 x_1}{2\pi x_2 x_3}} - 1\right)}$ |
| **III.7.38** | $\frac{4\pi x_0 x_1}{x_2}$ | III.8.54 | $\sin^2\left(\frac{2\pi x_0 x_1}{x_2}\right)$ |
| test 1 | $\frac{x_0^2 x_1^2 x_2^2 x_3^2 x_4^2}{16 x_5^2 \sin^4\left(\frac{x_6}{2}\right)}$ | test 2 | $\frac{x_0 x_1 \left(\sqrt{1 + \frac{2 x_2^2 x_3}{x_0 x_1^2}} \cos(x_4 - x_5) + 1\right)}{x_2^2}$ |
| test 3 | $\frac{x_0 \cdot (1 - x_1^2)}{x_1 \cos(x_2 - x_3) + 1}$ | test 4 | $\sqrt{2} \sqrt{\frac{x_1 - x_2 - \frac{x_3^2}{2 x_0 x_4^2}}{x_0}}$ |
| test 5 | $\frac{2\pi x_0^{\frac{3}{2}}}{\sqrt{x_1 (x_2 + x_3)}}$ | test 6 | $\sqrt{\frac{2 x_0^2 x_1^2 x_6}{x_2 x_3^2 x_4^2 x_5^4} + 1}$ |
| test 7 | $\sqrt{\frac{8\pi x_0 x_1}{3} - \frac{x_2 x_3^2}{x_4^2}}$ | test 8 | $\frac{x_0}{\frac{x_0 \cdot (1 - \cos(x_3))}{x_1 x_2^2} + 1}$ |
| test 9 | $-\frac{32 x_0^4 x_2^2 x_3^2 (x_2 + x_3)}{5 x_1^5 x_4^5}$ | test 11 | $\frac{4 x_0 \sin^2\left(\frac{x_1}{2}\right) \sin^2\left(\frac{x_2 x_3}{2}\right)}{x_1^2 \sin^2\left(\frac{x_2}{2}\right)}$ |
| test 12 | $\frac{x_0 \left(-\frac{x_0 x_1^3 x_3}{(x_1^2 - x_3^2)^2} + 4\pi x_2 x_3 x_4\right)}{4\pi x_1^2 x_4}$ | test 13 | $\frac{x_0}{4\pi x_4 \sqrt{x_1^2 - 2 x_1 x_2 \cos(x_3) + x_2^2}}$ |
| test 14 | $x_0 \left(-x_2 + \frac{x_3^3 (x_4 - 1)}{x_2^2 (x_4 + 2)}\right) \cos(x_1)$ | test 15 | $\frac{x_2 \sqrt{1 - \frac{x_1^2}{x_0^2}}}{1 + \frac{x_1 \cos(x_3)}{x_0}}$ |
| test 16 | $x_3 x_5 + \sqrt{x_0^2 x_1^4 + x_1^2 (x_2 - x_3 x_4)^2}$ | test 17 | $\frac{x_0^2 x_1^2 x_4^2 \cdot \left(1 + \frac{x_4 x_5}{x_3}\right) + x_2^2}{2 x_0}$ |
| test 18 | $\frac{3\left(\frac{x_1 x_4^2}{x_2^2} + x_3^2\right)}{8\pi x_0}$ | test 19 | $-\frac{\frac{x_1 x_5^4}{x_2^2} + x_3^2 x_5^2 \cdot (1 - 2 x_4)}{8\pi x_0}$ |

Table 11: **Formulas from the Feynman dataset.** The table lists 119 formulas from the Feynman dataset, derived from Feynman's lectures. Bold indicates formulas successfully recovered by the TF-16 model configuration. An asterisk (*) signifies recovery via generalization, where the formula was not present in the pre-training data.

Table 10: **Inference speed comparison results.** Comparison of the inference speed of the transformer-based symbolic regression approach with state-of-the-art search-based methods. The average inference time for a single expression is 2.3 orders of magnitude faster when inferred on an Intel(R) Xeon(R) Gold 6226R CPU @ 2.90GHz and 3 orders of magnitude faster when inferred on a single NVIDIA RTX A6000 GPU. The average times are calculated for a beam size of $k = 10$ over all expressions of the Feynman dataset.

| Method | Avg. Time per Expression (s) |
| --- | --- |
| **Method** | |
| DSR | 20.05 |
| gplearn | 9.10 |
| Operon | 7.28 |
| PySR | 3.82 |
| UDFS | 61.10 |
| **CPU Inference** | |
| TF-1 | 0.32 |
| TF-2 | 0.42 |
| TF-4 | 0.59 |
| TF-8 | 0.91 |
| TF-16 | 1.58 |
| **GPU Inference** | |
| TF-1 | 0.06 |
| TF-2 | 0.08 |
| TF-4 | 0.11 |
| TF-8 | 0.17 |
| TF-16 | 0.32 |

## 2.4 Performance Analysis on SRBench

This section situates the best-performing model configuration from our ablations, TF-16, within the context of state-of-the-art symbolic regression methods on the SRBench benchmark. Figure 8 illustrates the average recovery rate and complexity of the generated symbolic formulas. The transformer-based TF-16 model ranks second, following the search-based approach PySR. Notably, the TF-16 model generates less complex and more interpretable formulas, a result attributed to the carefully designed pre-training data selection. Beyond the recovery rate, the SRBench test suite offers insights into other performance aspects. As suggested by Kahlmeyer et al. [2024], the Jaccard index can serve as a non-binary measure of recovery, indicating the proportion of correctly recovered formula components. The left panel of Figure 9 presents the results of this evaluation. The TF-16 model configuration achieves the second-highest Jaccard index, demonstrating a strong ability to recover correct formula components.

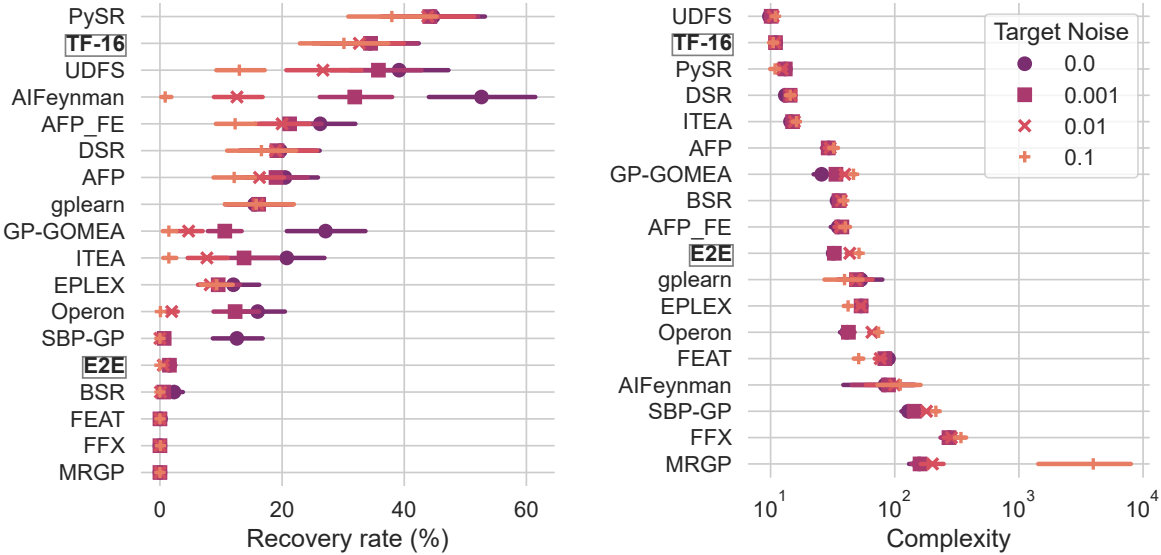

Figure 8: **Performance of state-of-the-art symbolic regression methods on SRBench.** Left: Recovery rate (percentage of correctly identified formulas) as a function of noise level for various methods. Right: Complexity of predicted formulas, measured as number of nodes in the expression tree [La Cava et al., 2021], for the same methods. Results are averaged over 10 runs.

Looking at model fit metrics such as the $R^2$-score, we can see that the TF-16 model configuration ranks lower than the E2E approach. This can be explained by the fact that we restrict the succinct expressions to contain at most 1 constant in order to obtain very compact expression representations. Other methods do not have this limitation and are therefore allowed to include an unlimited number of constants in the functions. This gives the generated candidates a much higher degree of freedom than the expressions generated by the TF-16 model configuration and explains their higher $R^2$-scores on the SRBench benchmark. The trade-off between the fit of the generated expression and its complexity is illustrated in Figure 10.

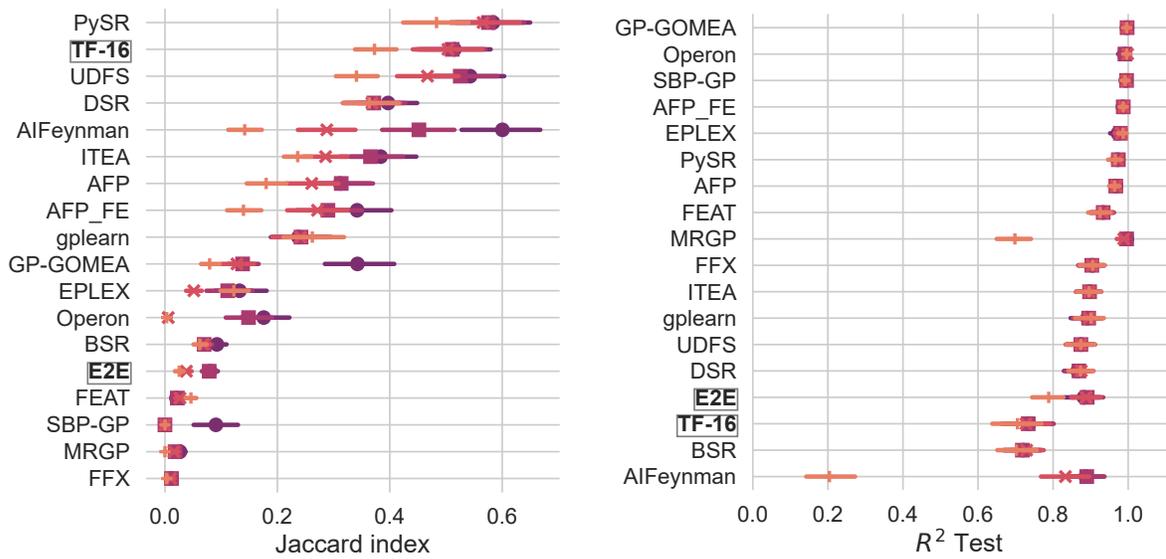

Figure 9: **Evaluation of formula similarity and goodness-of-fit on SRBench.** Left: Average Jaccard index, quantifying the overlap between the sets of variables and operators in the predicted and ground-truth formulas (higher is better). Right: Average R²-score, indicating the goodness-of-fit of the models to the SRBench data (higher is better). Each point represents a different method or configuration. Error bars show the deviation of the performance metrics (Jaccard index and R²) with varying noise levels (0 to 0.1). The reported values are averages over 10 runs.

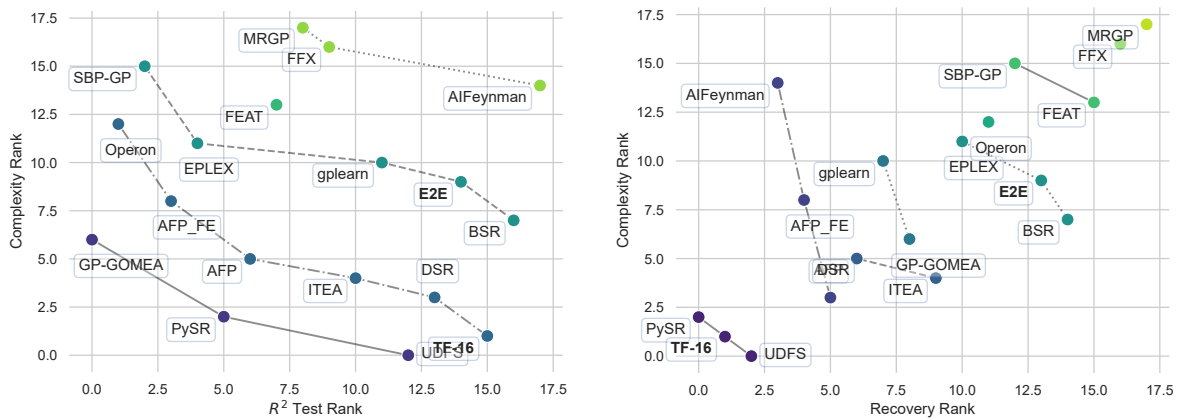

Figure 10: **Performance trade-offs on SRBench.** Left: Relationship between model complexity and goodness-of-fit ($R^2$) for various symbolic regression methods evaluated on the SRBench benchmark. Right: Relationship between model complexity and formula recovery rank for the same methods. Lower complexity and higher accuracy are desirable. Each point represents a different method or configuration. The Pareto front (best trade-off curve) is implied.

25

# 3 Robustness Analysis

In this section, we investigate the robustness of our pre-trained transformer models to variations in their training data. The central question is how stable these methods are when confronted with shifts in the sampling domain and distribution, factors that are critical during real-world application. We first detail our experimental setup, which consists of two distinct analyses. We then present a case study to illustrate model behavior under localized data perturbations, and conclude with a systematic analysis of performance as data is sampled increasingly far from the original pre-training domain.

## 3.1 Experimental Setup

To probe the robustness of our pre-trained models, we designed two complementary experiments focusing on their sensitivity to the sampling domain and distribution. For both experiments, we evaluate the two best-performing models from our ablation studies: the **Base** model and its **Standardized** variant, both pre-trained on enumerated expression DAGs with diverse sampling.

**Experiment 1: Case study on data perturbations.** Our first experiment is a case study designed to analyze model sensitivity to common data variations in the **training data** on a single, simple problem. We systematically apply four types of perturbations: (1) scaling the input domain, (2) shifting the input domain, (3) adding Gaussian noise to the output values, and (4) varying the sampling distribution itself. This allows for a granular analysis of how these factors individually impact model performance.

Let the initial input data be $X_{\text{init}}$ drawn from a domain $[a, b]$. For each perturbation, a parameter $\epsilon$ controls the magnitude of the change. Specifically, for an input data point $x_{\text{init}} \in X_{\text{init}}$ and its corresponding true output $y_{\text{true}}$, the transformations are defined as follows:

- **Input shift:** The data is shifted by a value proportional to the domain's range: $x = x_{\text{init}} + \epsilon(b-a)$.

- **Input scale:** The data is scaled by a value proportional to the domain's range: $x = x_{\text{init}} \cdot \epsilon(b-a)$

- **Output noise:** Zero-mean Gaussian noise is added to the output, with a standard deviation proportional to the signal's strength: $y = y_{\text{true}} + \mathcal{N}(0, \sigma^2)$, where $\sigma = \epsilon \cdot \text{RMS}(y_{\text{true}})$.

- **Distribution skew:** The sampling distribution is skewed using the transformation $x = a + (b-a)\left(\frac{x_{\text{init}}-a}{b-a}\right)^{\exp(-\epsilon)}$, which concentrates points towards the upper bound $b$ for $\epsilon > 0$ and the lower bound $a$ for $\epsilon < 0$.

**Experiment 2: Systematic Out-of-Domain Generalization.** Our second experiment systematically evaluates performance as a function of distance from the pre-training domain. The models were pre-trained on a hypercube of $[-10, 10]^d$. To test their OOPD capabilities, we sample training data from a series of concentric, non-overlapping cubical shells that extend progressively further from this initial hypercube. This setup allows us to precisely measure performance degradation as the data becomes increasingly OOPD.

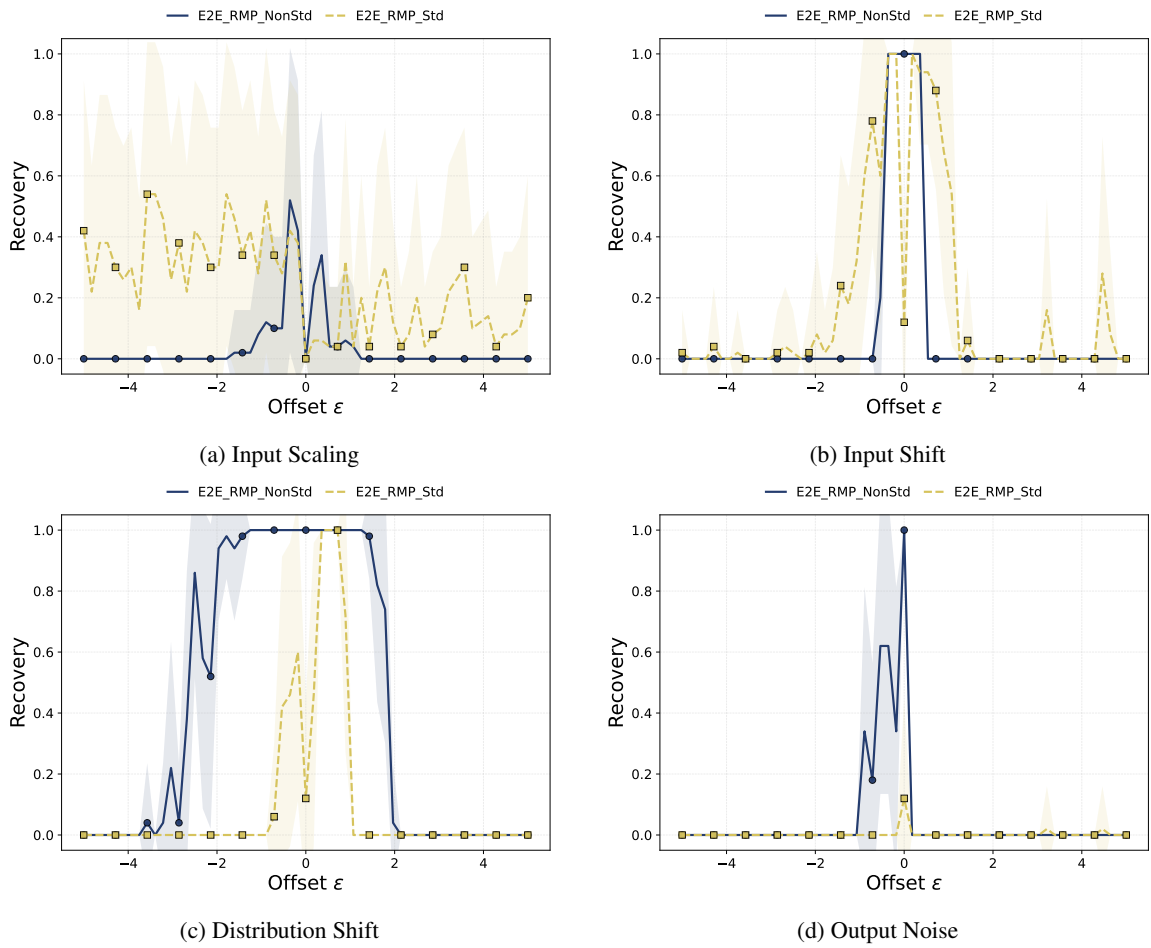

Figure 11: **Robustness analysis reveals a trade-off between standardized and non-standardized models.** The figure shows the recovery rate of two model variants—a non-standardized (**solid blue**) and a standardized (**dashed yellow**) version—on the simple problem $y = x^2$ under four types of training data perturbations. The parameter $\epsilon$ controls the magnitude of the perturbation, with $\epsilon = 0$ representing the original, unperturbed data. Lines represent the mean recovery over multiple trials, with shaded areas indicating the standard deviation. The results highlight a clear trade-off: **(b)** the standardized model is robust to **input shifts**, maintaining high recovery in a narrow window, but is extremely brittle to **(a) input scaling** and **(c) distribution shifts**. Conversely, the non-standardized model shows greater (though still limited) robustness to scaling and distribution shifts but fails completely on input shifts. **(d)** Both models are highly sensitive to **output noise**. This demonstrates that no single configuration is robust to all common data variations and that design choices like standardization do not solve this generalization problem.

## 3.2 Results

### 3.2.1 Case Study: Behavior Under Data Perturbations

The results of our case study, performed on the simple problem $y = x^2$, are presented in Figure 11. These tests are critical, as real-world data invariably contains variations against which models must be robust. Our findings reveal a fundamental trade-off between the standardized and non-standardized

27

models.

The standardized model demonstrates superior robustness to **input shifts** (Fig. 11b), achieving a high recovery rate within a narrow range around $\epsilon = 0$. However, this robustness does not extend to other perturbations. For **distribution shifts** (Fig. 11c) and **input scaling** (Fig. 11a), the standardized model is extremely brittle, with its performance collapsing immediately. In these cases, the non-standardized version shows a broader, albeit inconsistent, region of successful recovery. Furthermore, both models are highly sensitive to **output noise** (Fig. 11d), with recovery rates dropping strongly even for small noise levels.

These results lead to a crucial conclusion: even on a trivial problem, both pre-trained model variants exhibit significant fragility. This underscores a key limitation of current architectures and has important implications for their practical application. Before pre-trained models can be confidently deployed, they must be rigorously tested for stability under expected data variations. The stark contrast between this observed brittleness and the inherent robustness of non-pre-trained, search-based methods like PySR highlights a significant area for future research in making pre-trained models more reliable.

### 3.2.2 Systematic Out-of-Pre-Training Domain Shift Analysis

The results of our systematic OOPD shift experiment, presented in Figure 12, reveal the critical role of data standardization in increasing OOPD generalization. The performance of the non-standardized model collapses immediately for even the slightest shift away from its pre-training domain ($+10$), as seen across all metrics.

Specifically, the top panels show that for **symbolic recovery** (Fig. 12a, b), the non-standardized model's recovery and Jaccard index drop to near zero immediately. In stark contrast, the standardized model maintains a stable, albeit modest, performance, exhibiting a graceful degradation as the OOPD shift increases. A similar pattern emerges for **predictive accuracy** (Fig. 12c, d). The standardized model's ID and OOD $R^2$ scores remain consistent across the entire range of shifts, while the non-standardized model's accuracy plummets.

From these results, we draw a clear conclusion: data standardization is a crucial technique for enhancing the OOPD generalization of pre-trained models, preventing the immediate performance collapse seen in the non-standardized variant. However, it is essential to contextualize this improvement. The stabilized performance of the standardized model still remains significantly below that of non-pre-trained, search-based baselines, indicating that while standardization is beneficial, it is not a complete solution to the OOPD generalization challenge.

**Qualitative Analysis of Recovered Expressions.** To understand *why* the standardized model's performance degrades OOPD, we analyzed the specific symbolic expressions it successfully recovers in the IPD and OOPD settings. A clear pattern emerges: the set of expressions recovered OOPD is a strict subset of those recovered IPD. The failures are not random but are concentrated in specific functional classes. As shown in Table 12, the model consistently fails to recover expressions containing trigonometric functions (sin, cos) or complex rational structures in the OOPD setting. In contrast, the expressions it robustly recovers are predominantly composed of multiplication, division, and simple

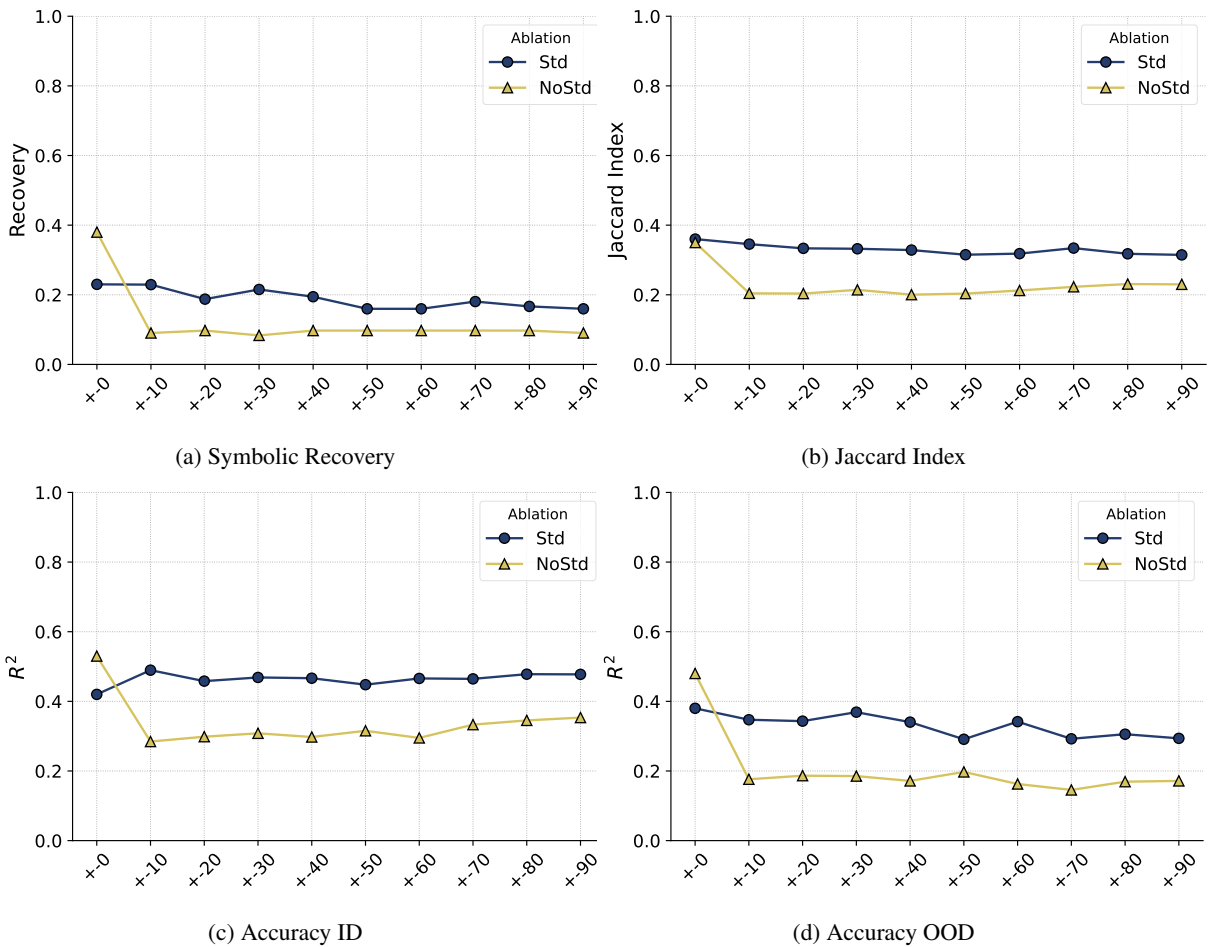

Figure 12: **Data standardization is important for improving recovery in OOPD scenarios.** The figure compares the performance of a model trained with data standardization (**Std**, solid blue) against one without (**NoStd**, dashed yellow). The x-axis represents the distance of the data-sampling region from the original pre-training hypercube, with '+0' being the in-domain case and '+10' to '+90' representing progressively distant, non-overlapping cubical shells. Performance is measured across four metrics: **(a)** Symbolic Recovery, **(b)** Jaccard Index, **(c)** in-domain accuracy (ID R²), and **(d)** out-of-domain accuracy (OOD R²). The non-standardized model exhibits a catastrophic performance collapse for any shift beyond its immediate pre-training domain ('+0'), with all metrics dropping to near-zero. In stark contrast, the standardized model maintains a stable, albeit modest, performance across the entire range of domain shifts. This demonstrates that while standardization does not lead to state-of-the-art performance, it improves performance in OOD scenarios.

powers.

This suggests that the model's ability to extrapolate is not uniform but is instead highly dependent on the structure of the target function. We hypothesize that this fragility stems from the model learning specific *functional forms in the standardized domain*, rather than the general rules of composition. For functions whose structure is relatively invariant to the affine transformation of standardization (e.g., monomials like $x^2$ becoming $(\sigma z + \mu)^2$, which is still a simple quadratic), the model can match

29

the familiar pattern seen during pre-training and successfully recover the expression. In contrast, for functions that are structurally sensitive to this transformation, the model fails. Periodic functions are a prime example: the in-domain $sin(z)$ becomes $sin(\sigma z + \mu)$ out-of-domain, a transformation involving phase and frequency shifts that drastically alters the function's behavior. The model's inability to handle this demonstrates that it has not learned to generalize the composition of the operator with the standardization mapping. Instead, it overfits to the specific patterns from its pre-training data, leading to a systematic failure on these more complex function classes in OOPD scenarios.

Table 12: **Examples of robustly vs. brittlely recovered expressions by the standardized model.** The table categorizes expressions based on whether they were recovered only IPD (Brittle) or in both IPD and OOPD settings (Robust). The model's OOPD failures are concentrated in trigonometric and complex rational functions.

| **Robustly Recovered Expressions (IPD & OOPD)** |
|---|
| `x_0*x_1*x_3/x_2` |
| `x_0*x_1**2/2` |
| `x_0/(4*pi*x_1**2)` |
| `sqrt(x_0*x_1/x_2)` |
| `-x_0*x_1 - x_1**2 + 2*x_1` |
| **Brittle Expressions (IPD Only)** |
| `-x_0*x_1*cos(x_2)` |
| `x_0*x_1*x_2*sin(x_3)` |
| `-sin(x_0) + 0.5*sin(x_0 - x_1)` |
| `1/(x_2/x_1 + 1/x_0)` |
| `x_2/(1 - x_1/x_0)` |

▶ We summarize the key outcomes of our robustness experiments as follows:

- **Pre-trained models exhibit significant fragility to data perturbations.** Our case study reveals a critical vulnerability of pre-trained models: even minor perturbations in input scale, shift, sampling distribution, or output noise can lead to a catastrophic drop in performance. The effects are observed despite using standardized data representations. This brittleness, which is not characteristic of non-pre-trained search-based methods, poses a major challenge for real-world applications where such data variations are common. This highlights a fundamental need for future research focused on developing more inherently robust pre-trained architectures.

- **Standardization improves OOPD robustness but does not solve the core generalization problem.** We find that data standardization is an important technique for mitigating catastrophic failure in OOPD scenarios, allowing the model to maintain a stable, gracefully degrading performance. However, our qualitative analysis reveals that this benefit is limited. The model's robustness is confined to expressions whose structure is relatively invariant under the affine transformation of standardization (e.g., generalized monomials), while it systematically fails on more sensitive structures like trigonometric functions. Therefore, while standardization is a valuable tool for stability, it is not a panacea for the fundamental challenge of OOPD generalization.

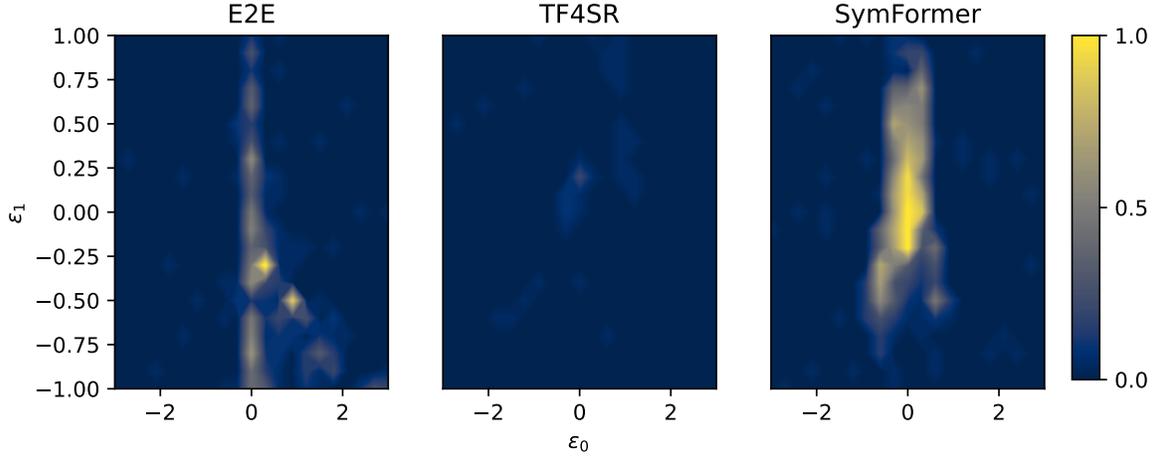

Figure 13: **Recovery performance is highly sensitive to shifts from the pre-training domain.** These heatmaps visualize the recovery rate of three transformer models as the training data distribution is parametrically shifted away from the original pre-training domain using shift parameters $\epsilon_0$ and $\epsilon_1$. The origin $(0,0)$ represents the in-pre-training domain, while points further away represent increasing out-of-domain shifts. Bright yellow indicates high recovery (1.0), and dark blue indicates low recovery (0.0). The results reveal a strong drop in performance for even minor shifts. E2E and SymFormer maintain high recovery only within a very narrow corridor around $\epsilon_0 = 0$, indicating extreme sensitivity to shifts along this axis. TF4SR fails to recover expressions across nearly the entire shift space. This provides strong visual evidence that these models learn patterns highly specific to their pre-training manifold and struggle to generalize beyond it.

### 3.3 Ablations on Pre-Trained Methods

**Pre-trained methods under input perturbations.** We visualize the behavior under input perturbations of three existing transformer-based symbolic regression methods [Kamienny et al., 2022, Lalande et al., 2023, Vastl et al., 2024]. We illustrate the behavior for the simple, exemplary expression $x(x+1)$. To do this, we generate training data for each method by transforming 100 equidistant base points $x_0 \in [0, 1]$. Specifically, these points are transformed using $10x_0 - 5$ for SymFormer [Vastl et al., 2024], $(x_0 - \bar{x}_0)/\sigma_{x_0}$ for E2E [Kamienny et al., 2022] (where $\bar{x}_0$ and $\sigma_{x_0}$ are the mean and standard deviation of $x_0$, respectively), and $100x_0/10$ for TF4SR [Lalande et al., 2023]. The models are then trained on these respective datasets to recover the function $f(x_0) = x(x+1)$. We measure recovery rates on training data generated from an interval $[a, b]$ with applied translational and scaling offsets $\varepsilon_0$ and $\varepsilon_1$. More specifically, we transform training data $x$ into $(x + \varepsilon_0(b-a))10^{\varepsilon_1}$. All reported recovery rates are averaged over 20 independent runs. The results are illustrated in Figure 13. We observe that the recovery performance of all approaches significantly degrades, eventually breaking down, as the training data is progressively shifted outside their pre-training domains.

We evaluate the accuracy of each approach using the $R^2$ score, with results presented in Figure 14. The E2E approach demonstrates consistently high accuracy, even as its training data shifts beyond the transformer's pre-training domain. This suggests that, for the specific formula tested, the E2E model can approximate the function $f(x_0) = x(x+1)$ effectively despite progressive domain shifts. This high performance is likely attributable to large number of free coefficients used in [Kamienny

31

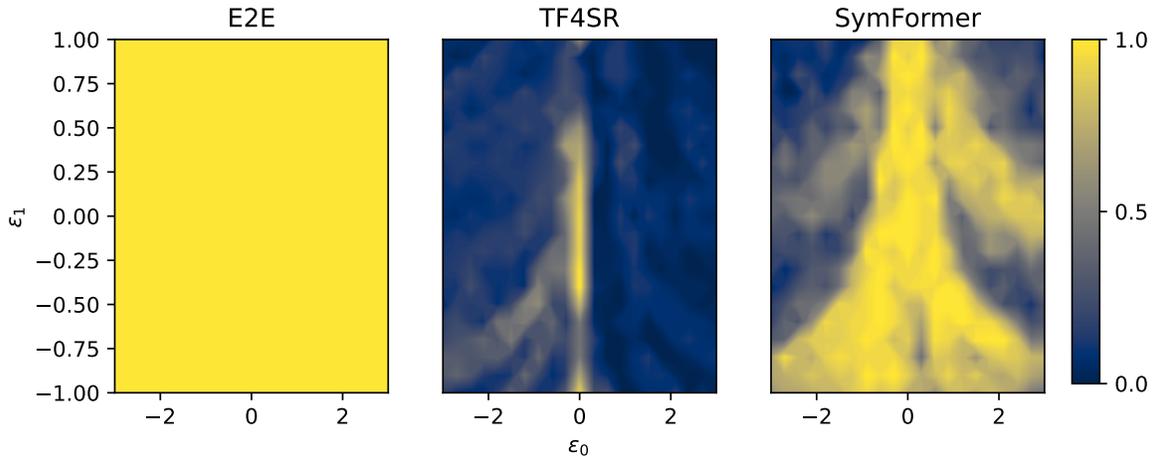

Figure 14: **Predictive accuracy (R²) reveals different model behavior under domain shift.** These heatmaps visualize the R² score (bright yellow is high) as the training data is parametrically shifted from the in-pre-training domain (origin) by parameters $\epsilon_0$ and $\epsilon_1$. The results show starkly different behaviors. **E2E** maintains near-perfect accuracy across the entire shift space, acting as a powerful universal curve fitter. Its ability to fit the data points numerically is completely decoupled from its ability to find the true symbolic expression (which fails, as shown in the previous figure). **In contrast**, the accuracy of **TF4SR** and **SymFormer** is highly sensitive to the domain shift, showing a more expected degradation pattern. This figure critically highlights that for some architectures, high accuracy can be a misleading metric of generalization capabilities, as it may mask a complete failure to recover the underlying generative model.

et al., 2022]. Given data from shifted domains, the subsequent parameter fitting via BFGS allows the model to fit the predicted skeleton to the data. In contrast, approaches lacking this subsequent fitting of parameters exhibit a consistent degradation in accuracy as their training data deviates further from their initial pre-training domain.

32



\end{document}